%% file: LangRecol_manuscript.tex
\documentclass[acmtog]{acmart}

\AtBeginDocument{%
  \providecommand\BibTeX{{%
    \normalfont B\kern-0.5em{\scshape i\kern-0.25em b}\kern-0.8em\TeX}}}

\setcopyright{acmlicensed}
\acmJournal{TOG}
\acmYear{2023} \acmVolume{42} \acmNumber{4} \acmArticle{101} \acmMonth{8} \acmPrice{15.00}\acmDOI{10.1145/3592111}

\definecolor{amber}{rgb}{1.0, 0.49, 0.0}
\definecolor{dodgerblue}{RGB}{30, 144, 255}
\definecolor{violet}{RGB}{238,130,238}
\definecolor{my_green}{RGB}{113,173,71}
\definecolor{my_blue}{RGB}{44,115,182}

\newcommand{\reffig}[1]{\textcolor{black}{Fig.~\ref{fig:#1}}}
\newcommand{\refsec}[1]{\textcolor{black}{Sec.~\ref{sec:#1}}}
\newcommand{\reftab}[1]{\textcolor{black}{Tab.~\ref{tab:#1}}}
\newcommand{\cherry}[1]{\textcolor{black}{#1}}
\newcommand{\zavier}[1]{\textcolor{black}{#1}}
\newcommand{\ryn}[1]{\textcolor{black}{#1}}
\newcommand{\rynn}[1]{\textcolor{black}{#1}}
\newcommand{\rynnn}[1]{\textcolor{black}{#1}}
\newcommand{\rynq}[1]{\textcolor{black}{#1}}

\newcommand{\wzw}[1]{\textcolor{black}{#1}}
\newcommand{\greentext}[1]{\textcolor{my_green}{#1}}
\newcommand{\bluetext}[1]{\textcolor{my_blue}{#1}}
\newcommand{\revise}[1]{\textcolor{black}{#1}}

\newcommand{\wzww}[1]{\textcolor{black}{#1}}

\newcommand{\eg}[1]{{\textit{e.g.,~}}}
\newcommand{\ie}[1]{{\textit{i.e.,~}}}
\newcommand{\etal}[1]{{\textit{et al.~}}}

\usepackage{enumitem}

\graphicspath{{./img}}
\DeclareGraphicsExtensions{.png,.jpg}

\begin{document}

\title{Language-based Photo Color Adjustment for Graphic Designs}

\author{Zhenwei Wang}
\authornote{Both authors contributed equally to this research.}
\email{zhenwwang2-c@my.cityu.edu.hk}
\orcid{0000-0003-0215-660X}
\affiliation{%
  \institution{City University of Hong Kong}
  \city{Hong Kong SAR}
  \country{China}
}

\author{Nanxuan Zhao}
\authornotemark[1]
\email{nanxuanzhao@gmail.com}
\orcid{0000-0002-4007-2776}
\affiliation{%
  \institution{Adobe Research}
  \city{California}
  \country{USA}
}

\author{Gerhard Hancke}
\email{gp.hancke@cityu.edu.hk}
\orcid{0000-0002-2388-3542}
\affiliation{%
  \institution{City University of Hong Kong}
  \city{Hong Kong SAR}
  \country{China}
  }

\author{Rynson W.H. Lau}
\email{rynson.lau@cityu.edu.hk}
\orcid{0000-0002-8957-8129}
\affiliation{%
  \institution{City University of Hong Kong}
  \city{Hong Kong SAR}
  \country{China}
}

\renewcommand{\shortauthors}{Trovato and Tobin, et al.}

\input{sections/abstract}



\input{sections/teaser}

\maketitle

\input{sections/introduction}

\input{sections/related_work}

\input{sections/overview}

\input{sections/synthetic_dataset}

\input{sections/color_prediction}

\input{sections/photo_recoloring}

\input{sections/evaluation}

\input{sections/application}

\input{sections/conclusion}

\bibliographystyle{ACM-Reference-Format}
\bibliography{LangRecol-base}


\end{document}

%% file: sections/abstract.tex
\begin{abstract}

\cherry{Adjusting \rynn{the photo color to associate with some design element\wzw{s}} is an essential way for \rynn{a graphic design to effectively deliver its message and make it} aesthetically pleasing. However, existing tools and previous works face a dilemma between the ease of use and level of expressiveness.
To this end, we introduce an interactive language-based approach for photo recoloring, which provides an intuitive system that can assist \rynn{both experts and novices on graphic design}.
Given a graphic design containing a photo that needs to be recolored, our model can predict the \wzww{source} \wzw{colors} and \rynn{the \wzw{target regions}, and then recolor the \wzw{target regions} with the \wzww{source} \wzw{colors}}
based on the given language-based instruction. The multi-granularity of the instruction allows diverse user intentions. The proposed novel task \rynn{faces several unique challenges,} including: \textit{1) color accuracy} for recoloring with \wzww{exactly} the same color from the target design element \rynnn{as} \wzww{specified by the user}; \textit{2) multi-granularity instructions} for parsing instructions correctly \wzww{to generate a specific result or multiple plausible ones}; and \textit{3) locality} for \rynn{recoloring}
in semantically meaningful local region\wzw{s} \wzww{to preserve original image semantics}. To \rynn{address} these challenges, we propose a model called \textit{LangRecol} with two main components: the language-based \wzww{source} color prediction module and the semantic-palette-based photo recoloring module. We also introduce an approach for generating a synthetic graphic design dataset with instructions to enable model training. We evaluate our model via extensive experiments and user \wzw{studies}. We also \rynn{discuss} several practical applications, showing the effectiveness and \ryn{practicality} of our approach. \wzww{Please find the code and data at \url{https://zhenwwang.github.io/langrecol}.}}
\end{abstract}

%% file: sections/teaser.tex
\begin{teaserfigure}
  \includegraphics[width=\textwidth]{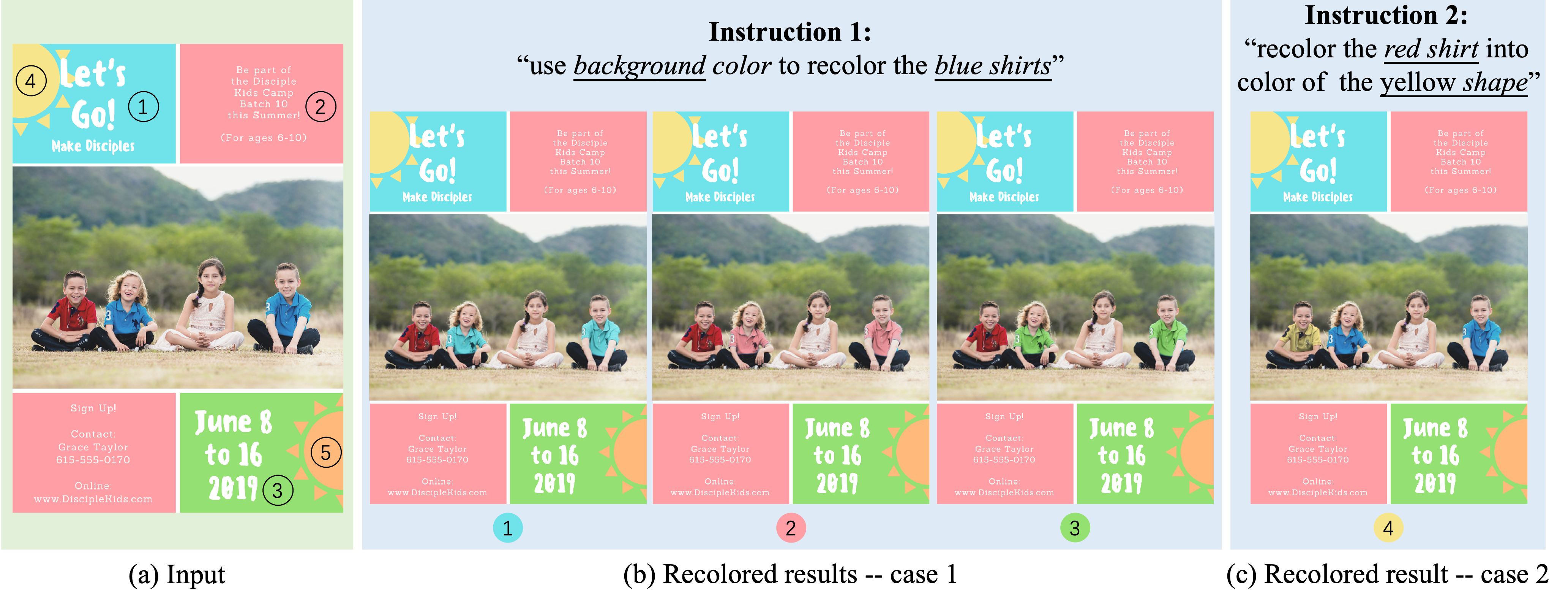}
  \caption{
 Language-based photo recoloring of a graphic design. Given a graphic design containing an inserted photo (a), our model recolors the photo automatically \ryn{according to the given language-based instruction}. To facilitate the expression of various user intentions, our model supports multi-granularity instructions for describing the \wzww{source} color\wzw{(s)} \ryn{(selected from the design elements) and the region(s) to be modified (selected from the photo)}. For example, the user may provide a coarse-grained instruction (``background'' in (b)) to refer to multiple \wzww{source} colors or a fine-grained instruction (``yellow shape'' in (c), \ie, using the yellow shape at the top left) to specify a \wzww{source} color. For visualization, we highlight the \wzww{source} \rynn{colors predicted by our model at the bottom (colors inside the circles)}.
  }
  \Description{Two examples of language-based photo recoloring for graphic designs}
  \label{fig:teaser}
\end{teaserfigure}

%% file: sections/introduction.tex
\section{Introduction}
Graphic designs (\eg, posters, webpages, slides and advertisements) have become a prevailing communication tool nowadays.
As \rynn{photos} play an essential role in graphic designs, \rynn{they} are often recolored to associate and harmonize
with other elements (\eg, text, background and shape) in the graphic designs in real applications~\cite{jorda2009brand, CreatePosterForBrand,huang2018automatic,PosterDesignGuide2020}. This helps the graphic \rynn{design deliver the message effectively, engage viewers and evoke} emotions.
Prior works have been conducted to help automate this task~\cite{zhao2021selective,cohen2006color,nguyen2017group,kim2018image}. 

However, when used by novices, existing methods and commercial software~(\eg~Photoshop, Affinity Photo, and GIMP)
often face a dilemma between the ease of use and level of expressiveness. Commercial software allows users to indicate arbitrary color modifications, but \wzww{requires} them to have design knowledge and rich experience. The work of Cohen-Or \etal~\shortcite{cohen2006color} allows users to adjust the image color to match with the color of the context elements through different harmonic schemes, but \rynn{it requires users to understand these schemes and it is often} difficult for them to control the color change within the specified local regions. 
The framework from Nguyen \etal~\shortcite{nguyen2017group} allows users to recolor a group of images based on the theme color of a design (\eg~brochure), but the color can be changed in undesirable ways (\eg~skin tone turns green) without considering the semantic contents of the images.
Methods such as Kim and Suk~\shortcite{kim2018image} and Zhao \etal~\shortcite{zhao2021selective} allow users to recolor a photo with \wzww{source} colors from the design by a single click, but these methods can only generate deterministic results that may not be the desired outcome. 

In this work, we aim to seek an interactive approach that is user friendly and has a broad range of expressiveness. We follow the basic setting of existing works~\cite{zhao2021selective, kim2018image}
by recoloring the target local region(s) in the photo with a \wzww{source} color extracted from the design.
However, we take advantage of the recent success of language-based interactions used in various \wzww{computer vision} tasks~\revise{\cite{zou2019language,chen2018language,ma2018language,jiang2021talk,Weng2022lcode,bahng2018coloring, luddecke2022image}}, and present a \wzww{language-based} system for photo color adjustment in the context of graphic designs (\reffig{teaser}). This system interaction comes natural to people and is intuitive to use.
It can also be combined with voice input and \wzww{is effective to the literacy education for children~\cite{zou2019language}.}
The system allows users to simply state their goals via concise verbal terms, without the need to learn a new interface or hunt through menus~\cite{woolfe2007natural,laput2013pixeltone, zou2019language}.

Although there are \rynn{some} works on language-based image editing and colorization,
designing such a system \rynn{specifically} for our task has several unique challenges.
1) \textbf{Color Accuracy:}
Rather than recoloring using an arbitrary color, \cherry{the \wzww{source} color should be exactly the same as the one \rynn{obtained from the design itself, as specified by the user}}.
Existing works~\cite{liu2020open,li2020manigan,zou2019language}
\ryn{that rely on specifying a vague color attribute (\eg~``blue'') with a wide range of possible mapped values may not be accurate enough}. How to interpret the instruction to extract the \wzww{source} color\wzw{s} from \rynn{the correct element\wzw{s} of the graphic design}
may not be straightforward.
2) \textbf{Multi-granularity Instructions:} 
To allow users \rynn{to express their objectives in a diverse way}, the input language-based instruction should be in multi-granularity, \ie~from an \wzw{obscure} one (\eg~\zavier{``text''})
to a more specific one (\eg~\zavier{``orange text'' or ``subtitle''}). The model needs to parse different instructions precisely, while some instructions may lead to multiple plausible results.
3) \textbf{Locality:}
\ryn{To preserve the image semantics, 
the color editing operation should be limited to} local regions. The model needs to understand image content semantically and propagate the \wzww{source} color into the local region\wzw{s} naturally to produce \wzw{a} high-quality result.

To tackle the above three challenges, we present a novel language-based recoloring framework, called \textit{LangRecol}, for \ryn{photo color adjustment} in the context of graphic \rynn{designs}. Given a graphic design with an inserted photo and a language-based instruction, our LangRecol system automatically adjusts the color of the photo following the instruction. The language-based instruction generally contains two parts, one for indicating \ryn{where to obtain the \wzww{source} color\wzw{s}} from the design elements, and the other for indicating \ryn{where the \wzw{target} local region\wzw{s} are in the photo for color adjustment}. The key idea of our framework is to take the language-based instruction as a bridge to model the relationship between graphic designs and the inserted photos in order to solve two main problems: predicting the specified \wzww{source} \wzw{color(s)} from the multi-granularity instruction by parsing the design correctly, and recoloring the target local region(s) of the photo semantically using the \zavier{predicted} \wzww{source} \wzw{color(s)}.

\cherry{\rynn{To solve} the first problem, w}e design a multi-task method to jointly conduct granularity recognition and \wzww{source} color prediction. The granularity recognition branch aims to recognize the granularity of the instruction to associate it with different types of target design element. \cherry{As the color distributions of different types of element (\eg~text and background \rynnn{colors}) are different, our \wzww{source} color prediction branch is customized based on the type of target element.}
To facilitate model training, we propose an approach to synthesize a graphic design dataset following some basic graphic design principles \cite{PosterDesignGuide2020,EffectivePosterDesign,MakeFlyerStandsOut} and  knowledge obtained from real-world graphic designs.
\cherry{\rynn{To solve} the second problem, w}e design a semantic-palette-based method to constrain the color editing step to be within semantically meaningful local regions. 
To do this, we first predict an initial region mask of the user specified object(s) by parsing the input instruction with the image content.
We then convert the initial region mask to refined coherent soft \wzw{semantic} color layers by leveraging both semantic features \cite{aksoy2018semantic} and color features \cite{tan2018efficient,wang2019improved}. 
Finally, we obtain the recoloring results by changing the color of the \zavier{target} layers to the predicted \wzww{source} colors. 

We evaluate the effectiveness of our method via extensive qualitative and quantitative experiments and user \wzw{studies}, \rynn{which show} that our pipeline \rynn{outperforms} existing automatic image color editing and language-based image color editing approaches. In summary, our main contributions are:
\begin{itemize}[leftmargin=*]
\item We design a novel tool called, \textit{LangRecol}, for photo color adjustment of graphic designs according to the multi-granularity language-based instruction.
\item We propose a multi-task model for parsing graphic design elements and understanding multi-granularity instructions through granularity recognition, \cherry{while predicting accurate \wzww{source} colors.}
\item We introduce a semantic-palette-based approach for language-based photo recoloring, predicting the target regions in a coarse-to-fine manner
and generating high-fidelity recolored results.
\item We develop a method to synthesize a plausible graphic design dataset based on real-world design knowledge and principles, to enable model training.
\end{itemize}

%% file: sections/related_work.tex
\section{Related Work}
To the best of our knowledge, the work \rynn{in}~\cite{zhao2021selective} has the most similar context with ours. While their goal is to generate deterministic recoloring results based on the user given colors, our work aims to adjust the photo color based on the input language-based instruction. This previous work needs explicit color assignment and limits the results \rynn{to be} within the regions predicted by the model \ryn{without any user control}. Instead, our work allows users to specify arbitrary \wzw{target} region\wzw{(s)} and \wzww{source} color\wzw{(s)} through an instruction. The rest of this section reviews prior works that are most related to ours.

\textbf{Scribble-based Image Recoloring.}
Scribble-based approaches require users to draw color strokes to interactively provide local color hints for recoloring. The propagation from the local color strokes to all pixels is usually based on low-level similarity metrics by optimization \cite{levin2004colorization,huang2005adaptive,yatziv2006fast,li2008scribbleboost,luan2007natural,xu2009efficient} or learning \cite{endo2016deepprop}. Recently, deep learning approaches are introduced to support sparse color strokes \cite{zhang2017real} or to solve \ryn{color-bleeding problems} \cite{kim2021deep}. 
While these works can produce appealing results, they often rely on extensive strokes. Instead, our model can generate recoloring results with a single instruction, supporting editing on design collections.

\textbf{Palette-based Image Recoloring.}
Palette-based approaches recolor images by manipulating the extracted color palette, which involves two critical problems: palette extraction and palette-based image decomposition.
The palette of a given image can be extracted using k-means methods \cite{chang2015palette, zhang2017palette}, \ryn{convex hull simplification} \cite{tan2016decomposing,tan2018efficient} or a physically-based method \cite{aharoni2017pigment}. For image decomposition, the soft color layers with alpha channels can be extracted by optimization over the alpha blending model  \cite{tan2016decomposing}, additive color mixing models  \cite{aksoy2017unmixing,zhang2017palette} or geometry-based \rynn{method} \cite{tan2018efficient}. \wzww{Recent deep-learning-based approaches \cite{afifi2021histogan,wang2022palgan} utilize a color palette or histogram features to globally control the colors of images generated by Generative Adversarial Network (GAN) models \cite{goodfellow2014generative}. Afifi \etal~ \shortcite{afifi2021cams} uses a color palette to segment the image and apply color-aware multi-style transfer.}
Inspired by these methods, we propose a semantic-palette-based photo recoloring approach, taking both color continuity in the palette and semantic features into consideration, so that only the colors within the target region\wzw{(s)} specified by the instruction are adjusted properly.

\textbf{Example-based Image Recoloring.}
Example-based approaches require users to provide reference images to guide the recoloring process. Traditional methods \cite{reinhard2001color,chang2005example,bae2006two,hacohen2013optimizing,arbelot2017local} 
transfer colors from a reference image to a target image based on low-level feature correspondences, \rynn{without considering} high-level semantic relationships.
\rynn{Recent} deep-learning-based approaches \cite{luan2017deep,he2018deep} take the semantics of a scene into consideration when transferring the colors between image pairs.
As users only need to provide a reference image, these methods can reduce user efforts.
However, \rynn{as} they \cherry{recolor the whole image based on this reference, without maintaining the necessary color features of non-modifiable regions,} \rynn{they are not suitable for solving} our task.

\textbf{Language-based Image Recoloring.}
\wzww{Recent breakthroughs in NLP \cite{vaswani2017attention, brown2020language, radford2021learning} show its power as a promising interaction paradigm for image editing, allowing natural and accessible control by succinct sentences or phrases as instructions.} An increasing number of works relying on pretrained image generator and text encoder for general image editing tasks have \rynn{been proposed} \cite{xia2021tedigan, patashnik2021styleclip, nichol2021glide,ramesh2022hierarchical}, \rynn{with} high-quality results.
However, these works regenerate each pixel without \rynn{preserving} the original object structure, \rynn{and thus} cannot be used to address our problem. \revise{For image recoloring,} early works \cite{laput2013pixeltone, cheng2014imagespirit} take pre-defined instructions and semantic labels as inputs for rule-based control, which may not be easily adapted to practical scenarios.
Recently, some works \cite{zou2019language, chen2018language, cheng2020sequential} train \wzww{GAN models} with either paired data (\ie~\wzww{source} images, language instructions and recolored images) or unpaired data~\cite{zhu2017your, nam2018text, dong2017semantic}, but \rynn{are applicable to only a} specific domain, such as fashion, sketch, flower and bird.

\ryn{Closer to} our problem setting, there is another direction of works that focuses on dealing with images across different classes and editing the appearance-related attributes without changing the object structure. 
Li \etal~\shortcite{li2020manigan} propose to selectively manipulate description-relevant regions, taking a trade-off between reconstruction and editing. 
To achieve both high-quality images and effective editing at the same time, Liu \etal~\shortcite{liu2020open} recolor images by vector arithmetic over visual and textual features, 
and adopt a sample-specific optimization.
\cherry{Although these methods can semantically recolor parts of an image based on the input text description 
they often fail to constrain color propagation within local semantic regions. More recently, Khodadadeh \etal~ \shortcite{khodadadeh2021automatic} propose a GAN-based method to recolor images given predefined color and object tags, which can \rynn{constrain color changes to} a local region but suffers from the color bleeding problem with imperfect region mask\wzw{s}. Kawar \etal~ \shortcite{kawar2022imagic} propose a semantic image editing tool based on 
a pretrained text-to-image diffusion model \cite{saharia2022photorealistic}.
In addition, the \wzw{vague} color instructions used in all these works are not suitable to deal with our requirement of using the accurate \wzww{source} colors extracted from the design \wzw{itself for recoloring}.}

%% file: sections/overview.tex
\section{Overview}
\begin{figure*}[!t]
  \centering
  \includegraphics[width=\linewidth]{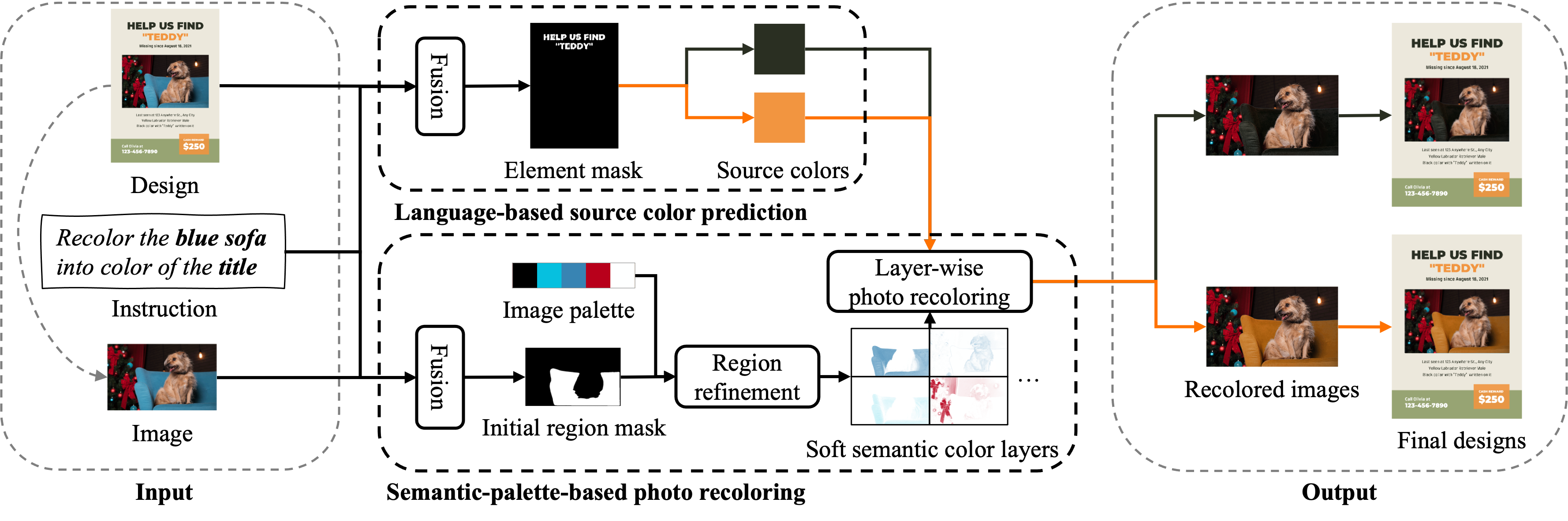}
  \caption{Overview of our pipeline. 
  \cherry{There are two key components in our pipeline: (1) \textbf{language-based \wzww{source} color prediction} aims to predict the \wzww{source} colors from the user-specified design elements (\ie~\rynn{``title''}); (2) \textbf{semantic-palette-based photo recoloring} aims to recolor \rynn{the user-specified local region\wzw{s} (\ie~``blue sofa'') in the photo with the  predicted \wzww{source} colors}.}
  }
  \Description{two stages: 1:color prediction; 2:photo recolor.}
  \label{fig:pipeline_overview}
\end{figure*}

\cherry{Our goal \rynn{in this work} is to construct a tool for recoloring \rynn{the} photos in graphic designs based on \rynn{the} given language-based instructions, which \rynn{is accessible to even} novices without any design experiences. We mainly focus on the single-page graphic design consisting of a small number of image, text and graphical elements, as it is widely used in our daily lives (\eg~poster, advertisement and flyer), and studied in recent works~\cite{o2014learning,zhao2021selective}.
\wzww{As design elements are not always available in vector format (\eg~a poster uses an inseparable image as background), we take pixel-format graphic designs as input.}
Given a \wzww{pixel-format} graphic design containing a photo that needs to be recolored, and a language-based instruction, our model parses the input instructions and extracts the \wzww{source} color(s) from the design elements while 
\wzw{locating and recoloring the target region(s) in the photo.}
}

\cherry{We show the pipeline of our model in \reffig{pipeline_overview}, which contains two key components: language-based \wzww{source} color prediction module and semantic-palette-based photo recoloring module.
As a graphic design usually contains
diverse elements \rynn{in the form of} a hierarchy (\eg~a text element may cover title, subtitle, and plain text), the input instruction should support multi-granularity
in order to identify the \wzww{source} color\wzw{(s)} properly.
For example, the user may use a coarse-grained description, such as ``background'', to refer to all colors of a specific type of element, as shown in~\reffig{teaser}(b); and a fine-grained color description, such as ``yellow shape'', to refer to a specific color (\ie~the color of the yellow shape, not the orange shape shown in \reffig{teaser}(c)).
The language-based \wzww{source} color prediction module aims to recognize the granularity and predict the \wzww{source} colors accordingly (see~\refsec{color_prediction}).
However, as high-quality graphic designs require professionals to create, collecting a large-scale dataset with designs, instructions and \wzww{source} colors for training is impractical. To mitigate this problem, we introduce an approach to synthesize graphic designs for training (see \refsec{dataset}). This allows us to obtain unlimited amount of data with ground truths easily.}

\cherry{To find the \rynn{target} region\wzw{s} in the image based on the instruction, there are two major requirements: semantically correct and locally compact. We take advantage of a large-scale dataset \zavier{PhraseCut \cite{wu2020phrasecut}} with semantic labels and annotated segmentation masks for learning. 
However, their annotations are not accurate and are rough near the boundaries. Recoloring with such imperfect masks \wzw{(\eg~a straightforward way is to predict a binary mask and apply the recoloring network from~\cite{zhao2021selective})} can generate \rynn{severe} and noticeable artifacts. Thus, we propose a coarse-to-fine method by taking this imperfect mask as an initial one, and then \rynn{refining} it by the \textit{semantic-palette-based region refinement} \wzw{process} with the help of \wzw{both deep semantic features and color continuity from palette extraction}. (see \zavier{\refsec{photo_recoloring}}).}

%% file: sections/synthetic_dataset.tex
\section{Synthetic Graphic Design Dataset with Instructions}
\label{sec:dataset}
\cherry{In this section, we introduce a method to synthesize \rynn{our} graphic design dataset, containing pairs of instructions and \wzww{source} color(s). Before introducing the model details, we first define the basic design elements with their hierarchy. We then discuss how we synthesize different design elements with multi-granularity instructions.}

\begin{figure}[!t]
  \centering
  \includegraphics[width=\linewidth]{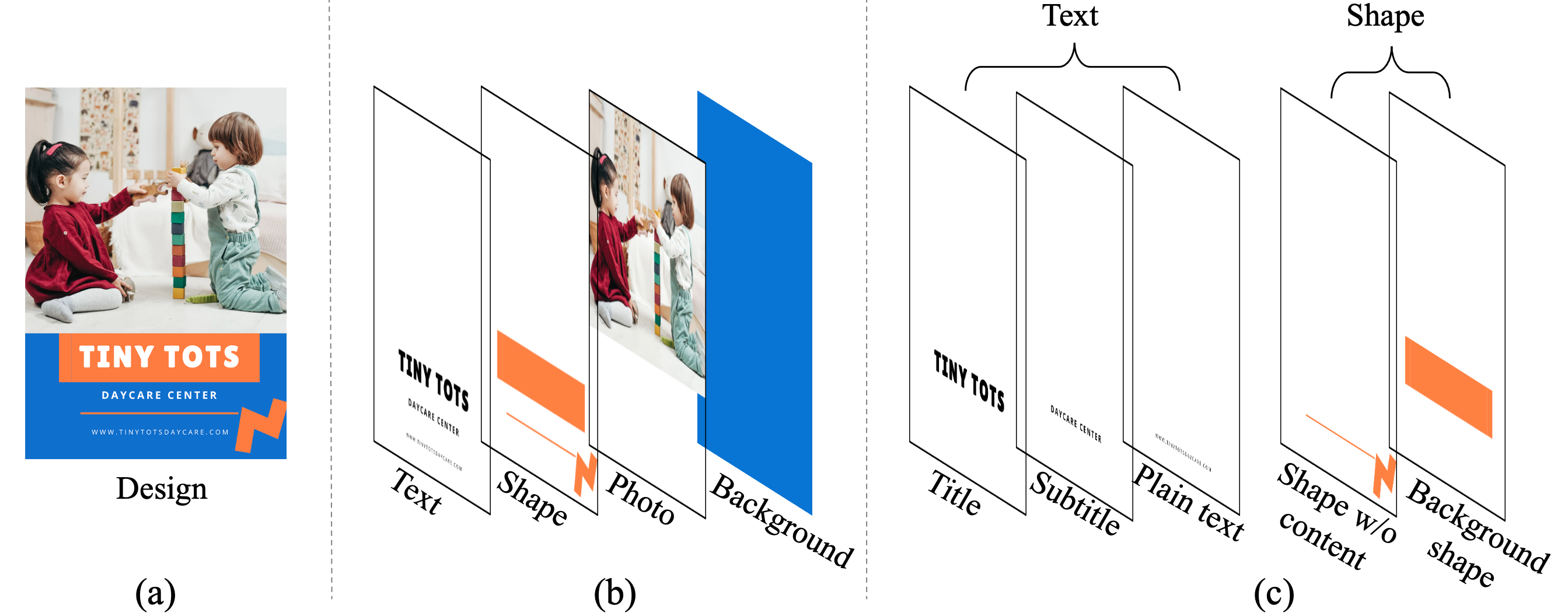}
  \caption{\cherry{\wzww{Hierarchy of eight types of basic design elements of a design in (a). The hierarchy consists of (b) three main categories (except photo), \ie~background, text, and shape, and (c) five fine-grained classes including title, subtitle, plain text, shape w/o content, and background shape.}}
   }
  \label{fig:element_term_examples}
\end{figure}

\cherry{\textbf{Basic Design Elements.} There are many types of elements in graphic designs. We start by collecting an initial list of frequently used design elements (except photos) from design books and blogs~\cite{PosterDesignGuide2020,EffectivePosterDesign,UltimateGuidetoFlyerDesign}. The initial list contains \zavier{16} types of design elements, covering three different categories: background, text (\eg~title, tagline, branding and call-to-action), and shape (\eg~logo, rectangle, polygon and line). To provide an accessible tool also for novices without any design experience, we remove elements that are not easily recognizable based on existing commercial tools (\eg~Canva\footnote{https://www.canva.com/templates/}) and a pilot study\footnote{We invited several graduate students from a local Computer Science department for this pilot study. We asked them whether they knew the meaning of each design element type and \rynn{whether they would use it in language-based instructions.}}. We obtain a final list of \wzww{eight} design elements with hierarchy, as shown in~\reffig{element_term_examples}.}

\subsection{Design Elements Generation and Layout}
Based on the above definition on design elements, \wzww{we synthesize a design $D \in \mathbb{R}^{H\times W\times 3}$ by combining several design elements with a photo image $I$.
For each synthetic design, we generate several language-based instructions $L=\{l_i\}$ that refer to different target elements} $E = \{e_i\} = \{e^{cls}_i, e^{att}_i\}$ with class labels (\eg~title) and color attributes (\eg~blue). For each \wzww{target} element $e_i$, we denote its RGB color \wzww{values} as  $c_i$ and binary mask as $m_i$.

\begin{figure}[!t]
  \centering
  \includegraphics[width=\linewidth]{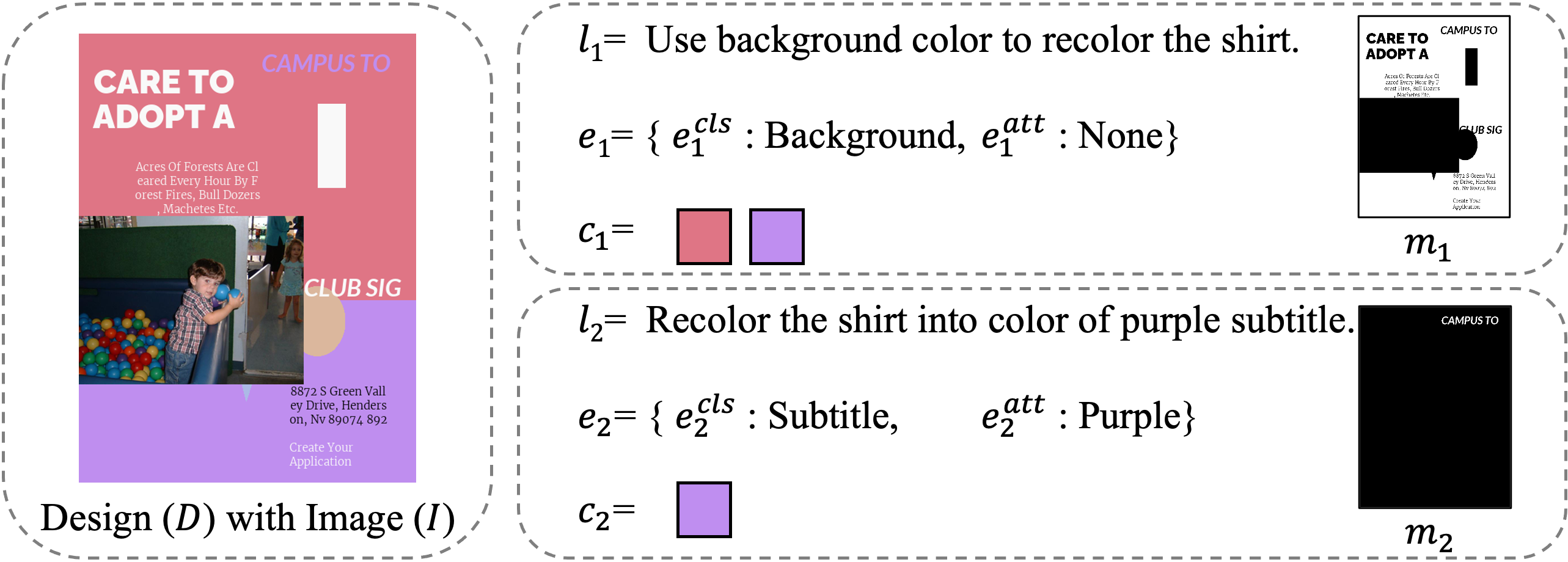}
  \caption{Example of our synthetic graphic design dataset. \wzww{For each design, we generate multiple language-based instructions referring to different target elements, with ground truth annotations (\eg~color and mask).}  
  }
  \label{fig:synthetic_design_example}
\end{figure}

Rather than randomly generating and arranging design elements, we design our synthesis method based on \revise{both graphic design principles (\eg~colors~\cite{PosterDesignGuide2020,EffectivePosterDesign,MakeFlyerStandsOut,UltimateGuidetoFlyerDesign} and fonts~\cite{BestGoogleFontToTry}), and design knowledge (\eg~layout and appearance) summarized from real-world graphic designs on Canva}. 
This can reduce the distribution gap between synthetic designs and the real ones, to better adapt our model to real design cases. An example of our synthetic graphic design dataset is shown in~\reffig{synthetic_design_example}. \wzw{Refer to the \wzw{supplementary material} for more details}.

\subsection{Multi-granularity Instruction Generation}
\cherry{The language-based instruction generally contains two parts: one for identifying the \wzww{source} color\wzw{(s)} from the design elements, and the other for \rynn{locating the target region\wzw{(s)}}
in the inserted image.}
\cherry{As for the first part, we categorize RGB color values into 11 color attributes (\ie~blue, brown, green, orange, pink, purple, red, yellow, black, grey, and white) followed by a linguistic study on basic color terms \cite{berlin1991basic}. We randomly assign either a specific element (\eg~``subtitle'', ``blue shape'') with its color as the ground truth, or a type of design element (\eg~``all text'') with multiple colors as the ground truth. We find that adding the color attribute of a design element to help indicate \rynn{the} specific \wzww{source} color is user-friendly and an effective way of mitigating ambiguity.}

\cherry{As for the second part, we use a large-scale dataset called PhraseCut~\cite{wu2020phrasecut}, containing phrases of object regions with annotated segmentation masks. The multi-granularity phrases has already contained, such as ``walking people'', ``black shirt'', and ``short deer''. Similarly, we can use the object phrase directly (\eg~``clothes'') or in combination with its original color (\eg~``blue clothes'') for narrowing it down to a specific target region. The corresponding masks can be directly taken as the ground truths. Finally, we use \rynn{the} predefined instruction patterns, such as ``recolor [region description] into [color description]'' or ``use [color description] to recolor [region description]'', to combine the region and color descriptions into a complete instruction. Note that the order of these two parts can be random to allow more free-form instructions during testing.} The conjunction words are randomly selected from a list of candidate words (\eg~use ``edit/change/adjust the color of'' to replace ``recolor''), to mimic diverse human expression habits. \wzww{Refer to the \wzww{supplementary material} for more details}.

%% file: sections/color_prediction.tex
\section{Our approach}

\begin{figure*}[!t]
  \centering
  \includegraphics[width=\linewidth]{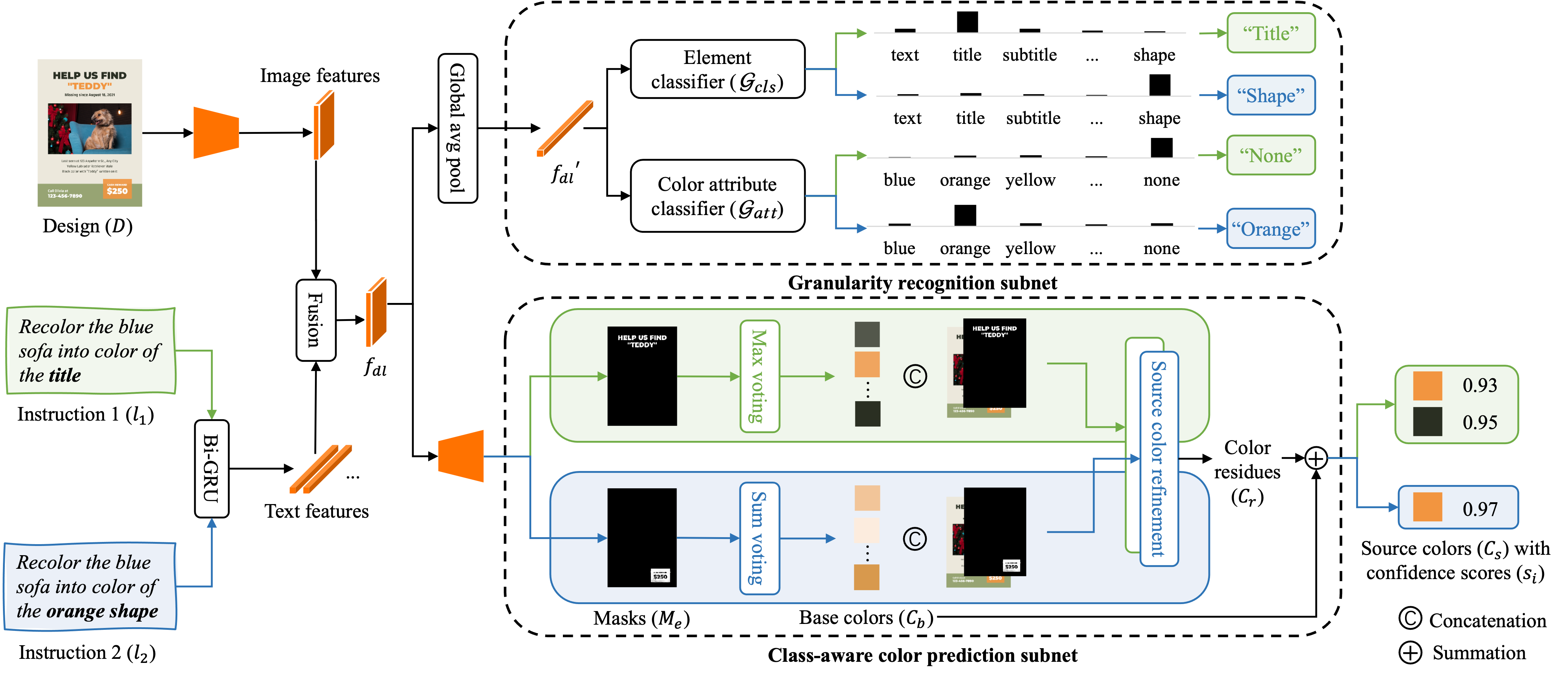}
  \caption{\cherry{Language-based \wzww{source} color prediction module. After fusing the features from the design and the instruction, the model conducts multiple tasks by recognizing the granularity of the instruction while predicting the \wzww{source} color(s) with confidence score(s). 
  \revise{\wzww{
  To better visualize the class-aware manner of our color prediction, we show two instructions referring to different types of target elements} (\ie, \greentext{green branch for text-based elements}, and \bluetext{blue branch for filled-color-based elements}). During inference, which branch the model \rynn{takes}
  depended on the class type predicted by the element classifier above.}
  }}
  \Description{granularity recognition module and class-specific color prediction}
  \label{fig:color_extraction}
\end{figure*}

\cherry{Given \rynn{a} design $D$ with \rynn{an inserted image} $I$, LangRecol generates \rynn{a} recolored image $I'$ \rynn{according to} the input language-based \wzw{instruction} $l$. As shown in~\reffig{pipeline_overview}, we first encode the design and the image separately using a ResNet101~\cite{zagoruyko2016wide}, and the instruction using Bi-GRU \cite{cho2014learning}, \rynn{to extract their} features. \rynn{These features are then} fused into two different groups before being sent to the two main modules of our framework. More specifically, the features derived from the design and the instruction are fused for language-based \wzww{source} color prediction, while the features derived from the image and the instruction are fused for semantic-palette-based photo recoloring.}

\cherry{We use an encoder fusion operation $\mathcal{F}$, which progressively inserts the features of \rynn{the} instruction into the last three res-blocks through a co-attention module. This has been validated as an effective operation for fusing multi-modal features in multi-level without increasing additional computational burden \cite{feng2021encoder}. We denote the fusion features from the design and the instruction as $f_{dl}$ and fusion features from the image and the instruction as $f_{il}$:
$f_{dl}=\mathcal{F}(ResNet(D),BiGRU(l)), \ \ f_{il}=\mathcal{F}(ResNet(I),BiGRU(l))$. }

\subsection{Language-based \wzww{Source} Color Prediction Module}
\label{sec:color_prediction}

\cherry{Given the fusion features $f_{dl}$, we aim to predict \rynn{the} \wzww{source} color(s) from specified design elements. As there may be more than one \wzww{source} color, we design the model to predict a set of colors $\{c_i\}$ with confidence scores $\{s_i\}\in [0,1]$, and we only \rynn{consider} the predicted colors with high confidence (\eg~$>0.5$) as \rynn{the \wzww{source} colors  \wzww{$C_s$} during test} time. Taking the challenges of multi-granularity in instructions and diverse appearances among design elements, we design a multi-task network for \wzww{source} color prediction, as shown in~\reffig{color_extraction}. It contains two sub-networks: the granularity recognition subnet, and the class-aware color prediction subnet. 
\revise{
The \rynn{element class type predicted by the granularity recognition subnet would help determine} the way of color prediction in the class-aware color prediction subnet.}}

\subsubsection{Granularity Recognition Subnet}
\label{sec:GRM}
\cherry{\rynn{This} subnet aims to \wzw{distinguish}
\rynn{the granularity of the instructions. As the type and color \wzw{attribute} of elements} determine the granularity of the instruction on the \wzww{source} color(s), we convert the problem into two classification tasks by recognizing the \wzw{element} class and color \wzw{attribute} of the \wzw{target} design element in the instruction. Note that the color attribute can be ``None'' as it is an optional choice. Formally, given the fusion features $f_{dl}$, we first pass \rynn{them} through a global average pooling layer (GAP) for flattening into \rynn{1D features} $\wzw{f_{dl}}^{\prime}=\operatorname{GAP}\left(\wzw{f_{dl}}\right)$, \wzw{and then send them} into two different classifiers ${\mathcal{G}_{cls}}$ and ${\mathcal{G}_{att}}$ \rynn{to obtain}:
\begin{equation}
\begin{array}{l}
p_{cls}={\mathcal{G}_{cls}}\left(\wzw{f_{dl}}^{\prime}\right), \ \ \ \ \ p_{att}={\mathcal{G}_{att}}\left(\wzw{f_{dl}}^{\prime}\right), \\
\end{array}
\end{equation}
where $p_{cls}$ and $p_{att}$ are the predicted probabilities of element class and color \wzw{attribute}, respectively.}

\subsubsection{Class-aware Color Prediction Subnet}
\label{sec:CSCP}

\cherry{Directly predicting \rynn{the} \wzww{source} colors in RGB values is not adequate as a slight change of RGB values can cause a huge difference in visual perception, which \rynn{contradicts}
to our goal of \rynn{extracting}
accurate color(s) from the design element. We thus design an approach by voting from the colors within the region\wzw{s} of the target element. We obtain the region mask \rynn{of element $e$, $M_e\in \mathbb{R}^{H\times W \times 1}\in [0,1]$,} through an element segmentation subnet, where a higher value indicates a higher probability that the RGB value of a pixel is the \wzww{source} color of the specified element.}

\cherry{Before conducting the voting, we observe that elements with distinct appearances can perform very differently. Text elements including \rynn{titles, subtitles and plain text have string-like appearances, often overlapped with background colors and shapes, and their colors tend to be} distributed in a sparse way. In contrast, \rynn{a background color or shape often covers a large area of pure color in a filled region, and the color is} distributed in a compact way. \rynn{As such}, we take different actions based on the class type: text-based elements, \rynn{or} filled-color-based elements. We introduce a class-aware voting strategy to obtain $k$ base colors $C_b \in \mathbb{R}^{k \times 3}$ from the input design.}

\cherry{We first take the unique colors within the design pixels whose values in \rynn{the} predicted masks are larger than an adaptive threshold \rynn{$v$} (\ie~the mean value of the element mask)
to form the initial candidates of \rynn{the} base colors. 
\rynn{As there may be many pixels with very similar colors \wzww{(\eg~noisy and gradient colors caused by image compression)} and taking all these colors as candidates can make training very slow, we filter out pixels of similar colors and assign only unique colors} into $b \times b \times b$ uniform histogram bins ($b = 6$) to form the final candidate sets. \rynn{We then} adopt max voting for text-based elements and sum voting for filled-color-based elements according to the predicted class type in~\refsec{GRM}. We regard the top-k unique colors ($k=10$) as the base colors. \revise{\rynn{More details can be found in the \wzw{supplementary material}}.}}

\cherry{We 
\wzw{predict the final \wzww{source} colors}
by computing a color residual for each of the base \rynn{colors}. We represent each base color as a pure color map $B_i \in \mathbb{R}^{H \times W \times 3}, i\in [1,k]$, and concatenate it with the element mask $M_e$ and input design $D$. Based on this concatenated vector, we build a \wzww{source} color refinement subnet to obtain the color residuals $C_r$ and confidence scores ${s_i}$. The final \wzww{source} colors $C_s$ are computed as $C_b+C_r$. Note that the \wzww{source} color refinement subnet is also class-aware, and that we train two separate versions for \wzw{text-based} elements and filled-color-based elements (\ie~same architecture \rynn{but} with different parameters). }

\subsubsection{Loss Functions}
\label{sec:loss_function}
\cherry{Our loss contains the following terms:}
    \cherry{\textit{Granularity Loss.} 
    We use \rynn{the} cross entropy loss to compute the predicted probabilities with the ground truth $\hat{p_{cls}},\hat{p_{att}}$ defined as:
\begin{equation}
\begin{array}{l}
L_{g}=\operatorname{CE}(\hat{p_{cls}}, p_{cls})+\operatorname{CE}(\hat{p_{att}}, p_{att}).
\end{array}
\end{equation}}
\cherry{\textit{Segmentation Loss.} We} adopt \rynn{the} binary cross entropy loss to measure the difference between the predicted element mask $M_e$ and the ground truth binary mask $\hat{M_e}$ as:
\begin{equation}
L_{m}=\hat{M_{e}} \log M_{e}+\left(1-\hat{M_{e}}\right) \log \left(1-M_{e}\right).
\end{equation}
\cherry{\textit{\wzww{Source} Color Loss.}} For \wzww{source} color prediction, we use a $L_2$ loss function to minimize the distance between the ground truth and predicted \wzww{source} colors, which is defined as:
\begin{equation}
\begin{array}{l}
L_{c}\left(\hat{C_s}, C_s\right)=\left\|\hat{C_s}-C_s\right\|_{2}^{2}.
\end{array}
\end{equation}
\cherry{\textit{Confidence Loss.} We adopt a binary cross entropy loss for \rynn{predicting the confidence scores of the} \wzww{source} colors. We take the confidences of ground truth \wzww{source} colors as $1$, and the others as $0$, and define the loss term as:
\begin{equation}
\begin{array}{l}
L_{s}=\hat{s} logs + (1 - \hat{s}) log(1 - s).
\end{array}
\end{equation}}
\cherry{In summary, the overall multi-task loss can be formulated as:
\begin{equation}
\begin{array}{l}
L=\lambda_{1} L_{g}
+\lambda_{2} L_{m}
+\lambda_{3} L_{c}
+\lambda_{4} L_{s},
\end{array}
\end{equation} where ${\lambda_i}$ are coefficients to balance the loss terms.}

%% file: sections/photo_recoloring.tex
\begin{figure*}[!t]
  \centering
  \includegraphics[width=\linewidth]{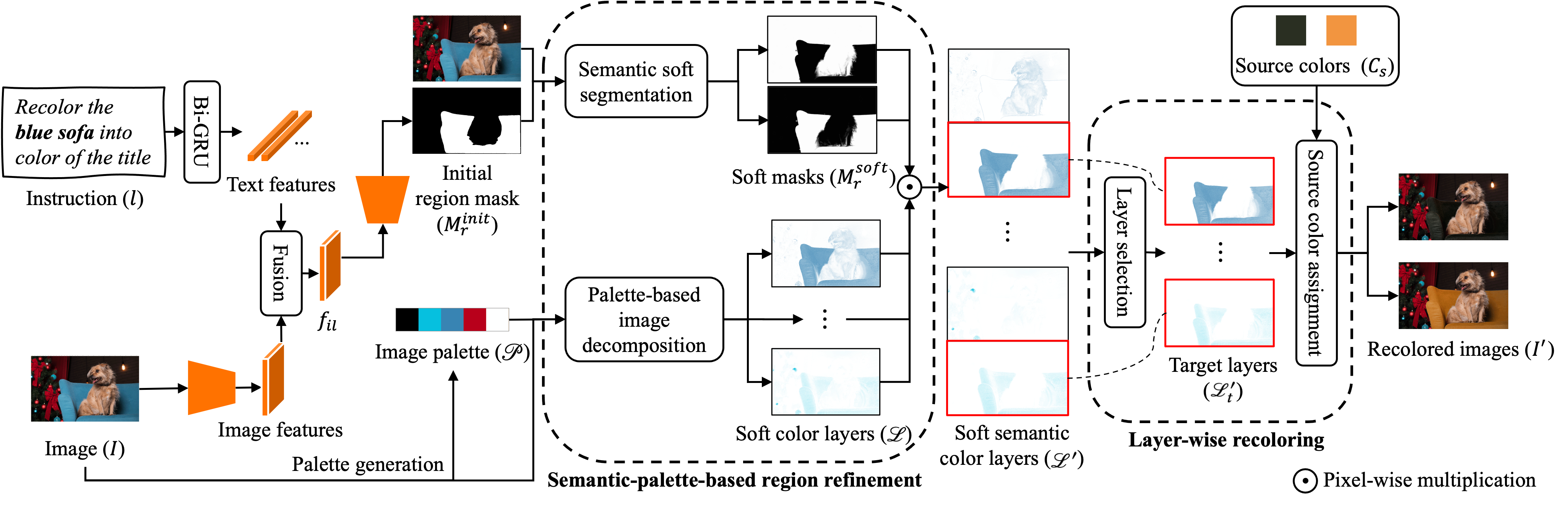}
  \caption{\cherry{Semantic-palette-based photo recoloring module. This module \wzww{first} predicts the target \wzw{region(s)} \wzww{in the form of soft semantic color layers} by taking advantages of both deep semantic features and color continuity from palette extraction. \wzww{It then performs photo recoloring in a layer-wise manner by assigning the source color(s) to each of the selected target layers (\ie~boxes marked in red).}}}
   \label{fig:photo_recoloring}
   
\end{figure*}

\subsection{Semantic-palette-based Photo Recoloring Module}
\label{sec:photo_recoloring}

\cherry{Our objective of this module is to predict \rynn{target} \wzw{region(s)} based on the fused features $f_{il}$, and recolor the image within the predicted \wzw{region(s)} to obtain a visually natural and pleasing result. \rynn{However, simply} relying on a deep neural network to learn semantics and predict the region for recoloring~\cite{zhao2021selective} can generate serious artifacts as we do not have large-scale well-annotated language-based segmentation dataset. Inspired by the recent success of palette-based recoloring methods~ \cite{chang2015palette,tan2016decomposing,tan2018efficient}, we take a hybrid way by combining the semantics-aware deep learning method and palette-based recoloring. The key idea is to mitigate the problem of imperfect region mask prediction by the network with color continuity of palette-based methods.
}

\cherry{As shown in~\reffig{photo_recoloring}, the module takes several steps: \rynn{(1)} generating initial region mask; \rynn{(2)} refining the initial region based on both semantic features from the deep network and soft color layers from the palette; \rynn{(3)} recoloring the image in a layer-wise manner. To generate the initial region mask $M_r^{init} \in \mathbb{R}^{H \times W \times 1}$ and acquire the semantic knowledge, we train a language-based image segmentation subnet, sharing the same network architecture as the one in~\refsec{CSCP}. The initial region mask is a hard one with binary values $M_r^{init} \in \{0,1\}$. We also obtain a color palette $\mathscr{P} \in \mathbb{R}^{N \times 3}$ using a geometry-based method~\cite{tan2016decomposing,tan2018efficient} for the inserted image. We then show how we refine the initial mask based on the extracted palette \rynn{and how we perform layer-wise \wzw{photo} recoloring}.}

\subsubsection{Semantic-palette-based Region Refinement}\label{sec:region_refinement}
\revise{\rynn{This step aims} to decompose the whole image into several soft semantic color layers, where each layer is a compact local region sharing the same semantics and color. 
We take a coarse to fine way to obtain the final layers.}
For semantics, we only distinguish between the target object (\rynn{referred to} as ``foreground'' in our context) and the other regions (\rynn{as} ``background''). Given the initial region mask $M_r^{init}$, we first compute the refined soft background and foreground region masks ${M}_{r}^{soft} = \{m_{f}, m_{b}\} \in [0,1]$ through a semantic soft segmentation method $S$~\cite{aksoy2018semantic}. This method leverages texture, color, and semantic cues for conducting image soft segmentation. We set the number of target segments to two and regard our initial region mask as the semantic feature vector for computing the semantic affinity term in $\mathcal{S}$. This can encourage the model to group pixels belonging to the user specified target region\wzw{(s)} together. \rynn{Formally},
\begin{equation}
\begin{array}{l}
{M}_{r}^{soft}=\{m_{f}, m_{b}\} = \mathcal{S}(M_r^{init}, I).\\
\end{array}
\end{equation}

The palette-based image decomposition method can generate a set of soft color layers $\mathscr{L} = \{\mathscr{l}_1,\mathscr{l}_2,...,\mathscr{l}_N\}, \mathscr{l}_i \in \mathbb{R}^{H \times W \times 4}$, each \rynn{corresponding} to a value in the color palette $\mathscr{P}$ with \zavier{an opacity value $\alpha^p \in [0,1]$ for each pixel. \rynn{The original image can then be restored by summing} these soft color layers $I=\sum_i \mathscr{l}_i$ and $ \sum_i{\alpha_i^p}=1$.}
Finally, we derive the soft semantic color layers $\mathscr{L}^{\prime}= \{\mathscr{l}'_1,\mathscr{l}'_2,...,\mathscr{l}'_Q\}$ by multiplying the soft color layers $\mathscr{L}$ and \wzw{refined soft region masks} $M_r^{soft}$ in a pairwise way (\ie~$Q=2N$), as shown in~\reffig{photo_recoloring}.

\subsubsection{Layer-wise Photo Recoloring}
\label{sec:layer-based_recoloring}

\cherry{With the benefit of linearly decomposing the input image as soft semantic color layers, we can directly recolor the image by changing the color of the target layers given the \wzww{source} color. 
\zavier{We select the target layers based on the foreground soft region mask $m_f$.}
Specifically, we compute region overlap rates $O = \{o_1,o_2,...,o_{\wzw{Q}}\}$ between the opacity value $\alpha_i$ of each \wzw{soft semantic color layer} and the foreground soft region mask $m_f$ as:
\begin{equation}
\begin{array}{l}
o_i = \revise{\frac{\sum_p(\alpha_i \odot m_f)}{\sum_p(\alpha_i)}},
\end{array}
\end{equation}
where $\sum_p(\cdot)$ denotes pixel-wise summation and $\odot$ denotes pixel-wise multiplication. As a \rynn{larger} $o_i$ indicates \wzw{a more dominant layer that contributes more color information} within the \rynn{target region\wzw{s}}, we select the top-n layers ($n=4$, empirically) \revise{as the target layers $\mathscr{L}^{\prime}_t= \{\mathscr{l}'_{t1},\mathscr{l}'_{t2},...,\mathscr{l}'_{tn}\}$, where $\mathscr{l}'_{t1}$ is the \wzw{most} dominant layer, and modify the color of these layers based on the \wzww{source} color $C_s$.}}

\revise{To preserve the original texture gradient in the target region\wzw{s}, instead of copying the \wzww{source} color to these layers directly, we follow the change of \wzw{\textit{lightness} (\ie~the $\mathbb{L}$-channel of CIE Lab color space, which is more aligned with human perception)} in the original region\wzw{s}. 
We define the original \wzw{\textit{lightness difference}} between the $i$-th target layer $\mathscr{l}'_{ti}$ and the \wzw{most} dominant layer $\mathscr{l}'_{t1}$ as: $\Delta{\mathbb{L}}_i = \mathbb{L}_{ti} - \mathbb{L}_{t1}$. After assigning the \wzww{source} color to each of the target layers, we alter the \rynq{\wzw{\textit{lightness}}} by adding back the above \wzw{\textit{lightness difference}} $\hat{\mathbb{L}}_i = \mathbb{L}_{C_s} + \Delta{\mathbb{L}}_i$.
By combining with unmodified layers,} we obtain the final recolored image.

%% file: sections/evaluation.tex
\section{Experiments}
\begin{figure*}[t]
  \centering
  \includegraphics[width=\linewidth]{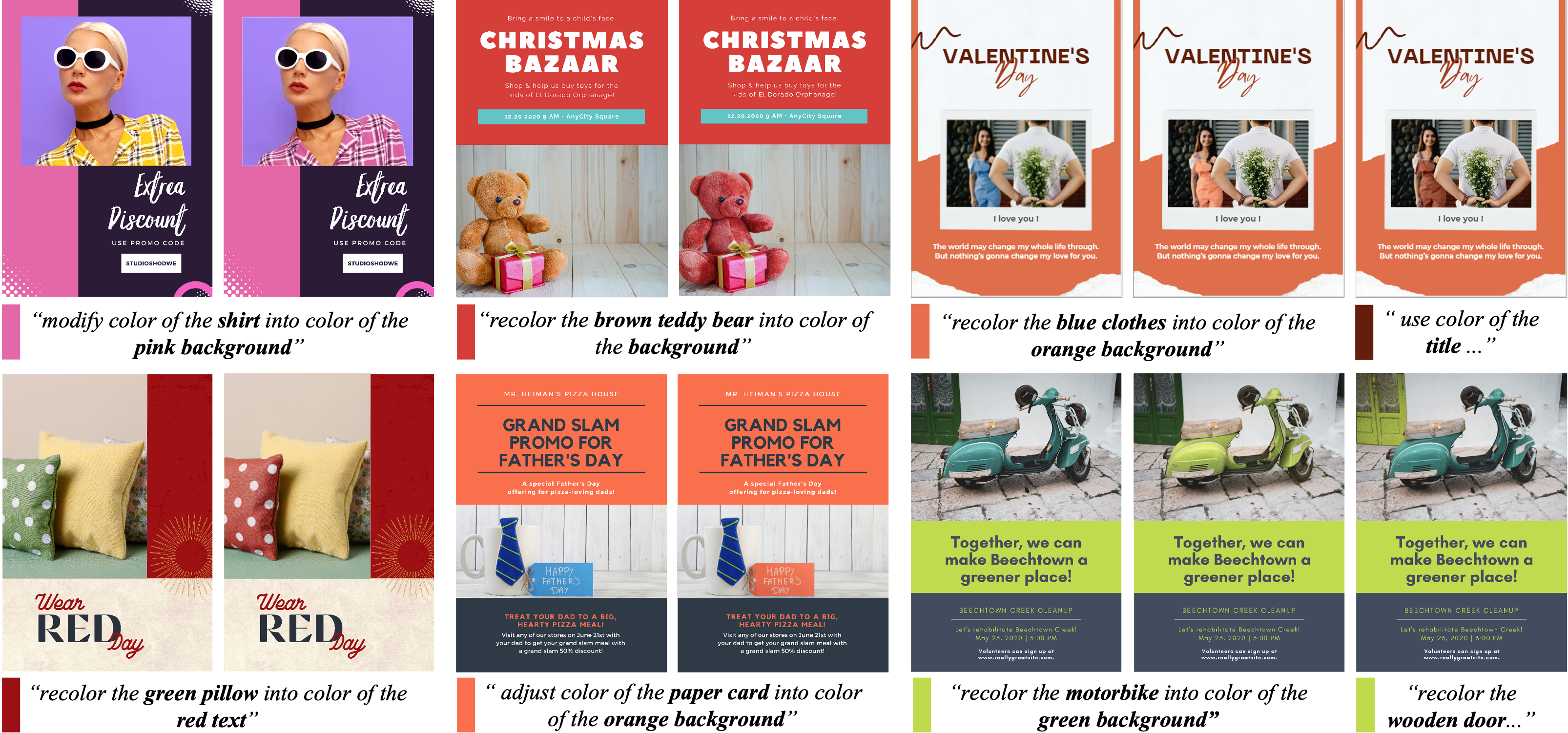}
  \caption{\cherry{Our \rynn{language-based photo recoloring results on graphic designs}. For each design case, we show the original design on the left and our result(s) on the right, with the predicted \wzww{source} color and the input instruction below. We highlight the words related to the \wzww{source} colors and \rynn{target} regions in bold.}
  }
  \label{fig:recolored_results}
\end{figure*}

\begin{figure}[t]
  \centering
  \includegraphics[width=\linewidth]{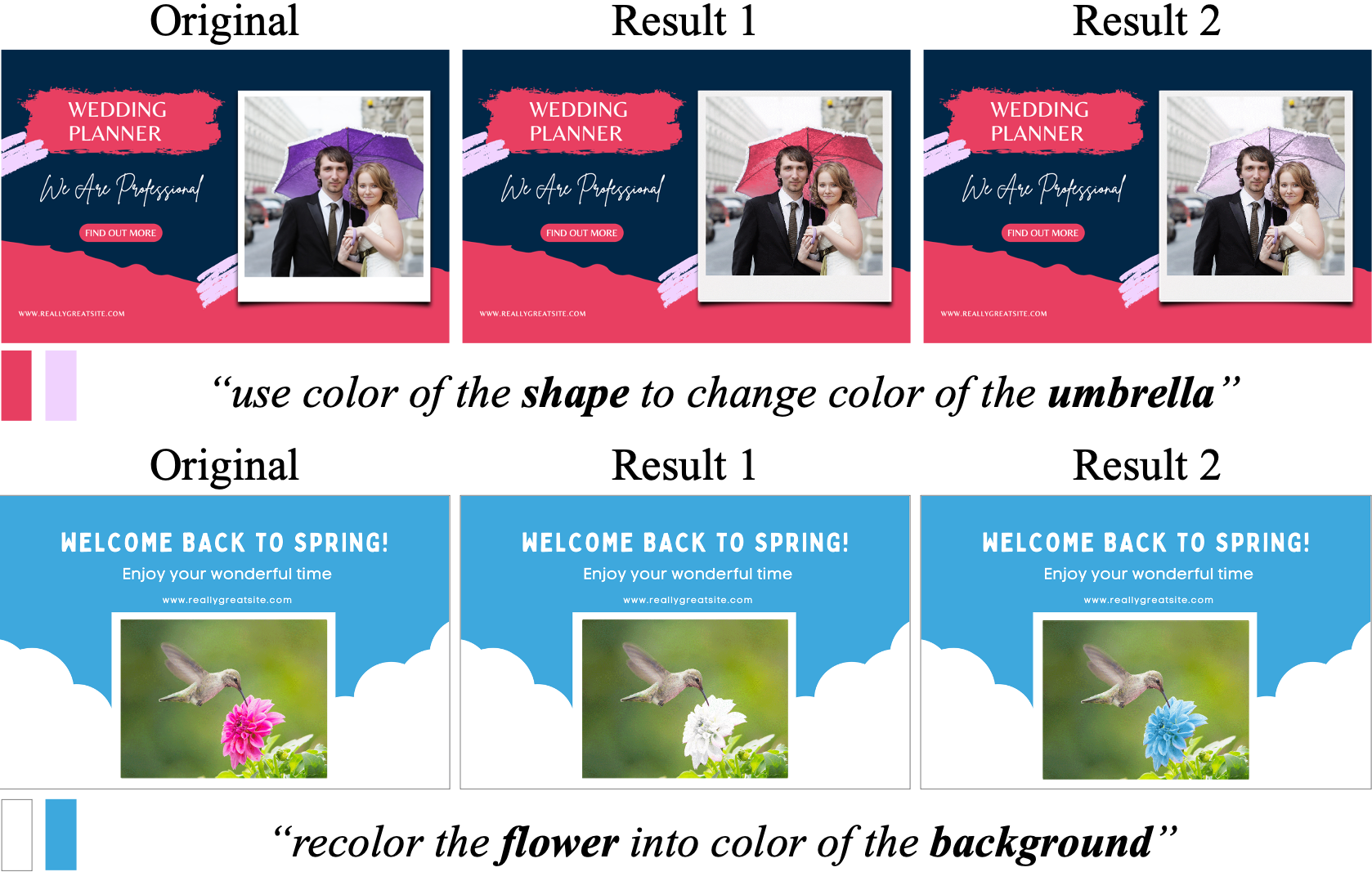}
  \caption{\cherry{Our \rynn{recoloring results} with coarse-granularity instructions. Left-bottom corner of each design case \rynn{shows} the predicted \wzww{source} colors.}
  }
  \label{fig:multi_color_recolored_results}
\end{figure}

\cherry{We show our \rynn{language-based \wzw{photo} recoloring results on} various \wzw{graphic designs with diverse appearances} in~\reffig{recolored_results}. 
\rynn{We can see that} our model can follow the instruction for recoloring, obtaining natural and pleasing results. The \rynn{designs also look} more visually harmonious after the photo color adjustment. In addition, our model allows users to specify different \wzww{source} colors and \rynn{target} regions through instructions to obtain multiple results (\eg~rightmost column of~\reffig{recolored_results}). As shown in~\reffig{teaser} and ~\reffig{multi_color_recolored_results}, if a coarse-granularity instruction (\eg~``color of the shape'') specifies multiple \wzww{source} colors, our model can also correctly generate multiple recoloring results. \rynn{For the rest} of this section, we first compare our results with those generated by \rynn{the} state-of-the-art methods and professional designers. We then conduct experiments to analyze the effect of different important design choices of our model.} 

\textbf{Implementation details.} We train the language-based \wzww{source} color prediction module (in~\refsec{color_prediction}) on the synthetic dataset (in~\refsec{dataset}) and the language-based image segmentation subnet (in~\refsec{photo_recoloring}) on the PhraseCut dataset. When training the language-based \wzww{source} color prediction module, we use a curriculum training strategy by pretraining the element segmentation subnet and the granularity recognition subnet first to avoid model collapse, and then train the whole model jointly. All the training and test processes are conducted on a PC with i7-10700 CPU and a single RTX3080 GPU. 
\rynn{More training details can be found in the \wzw{supplementary material}.}

\textbf{Test Set.}
\cherry{
We \rynn{have collected} 71 realistic graphic designs from Canva, and assign several diverse language-based instructions to each design \rynn{to obtain} a test set with 165 design-instruction pairs. These collected graphic designs cover a broad range of themes and styles, \wzw{such as} event posters, business advertisements, holiday flyers, etc. The inserted photos also span a wide range, from indoors to outdoors, from street to fashion, and from food, human to animal. As there are no ground truth recoloring results, we asked \wzww{three} professional designers to manually extract the \wzww{source} colors and recolor the inserted photos following the instructions. We \rynn{let them to use any tools that they preferred}, \eg~ Photoshop. \wzww{After two designers have finished all designs, the third designer chooses a preferred one in each design case as the ground-truth \wzww{source} colors and recoloring results.}}

\subsection{Comparison with Exiting Methods}
\subsubsection{Baselines} \revise{There \rynn{are no existing methods} supporting language-based photo recoloring in graphic \rynn{designs}. We thus \rynn{adapt} several state-of-the-art language-based image editing methods \rynn{in order for them to be applicable to our task:}}
\begin{itemize}[leftmargin=*]
    \item \revise{Four state-of-the-art text-guided image editing methods: Open-edit \cite{liu2020open}, ManiGAN \cite{li2020manigan}, RecolGAN \cite{khodadadeh2021automatic} and Imagic \cite{kawar2022imagic}, \rynn{which} support structure-preserved recoloring. 
    To fit our problem, we manually convert our instructions to the prompts in a format \rynn{suitable for these methods}. For example, we use ``blue sofa'' for ManiGAN, ``red sofa -> blue sofa'' for Open-edit, ``recolor flower to orange'' for RecolGAN, and ``a photo of a blue car'' for Imagic.}
    \item \cherry{A state-of-the-art work for photo recoloring in graphic design \cite{zhao2021selective} (GDRecolor). This method directly predicts modifiable \rynn{regions based on a specified color, and recolors these regions} in an inpainting manner. As this method cannot process language-based input, we directly take the \wzww{source} colors predicted by our model to obtain its results.}
\end{itemize}

\subsubsection{Qualitative Results}
\begin{figure*}[t]
  \centering
  \includegraphics[width=\linewidth]{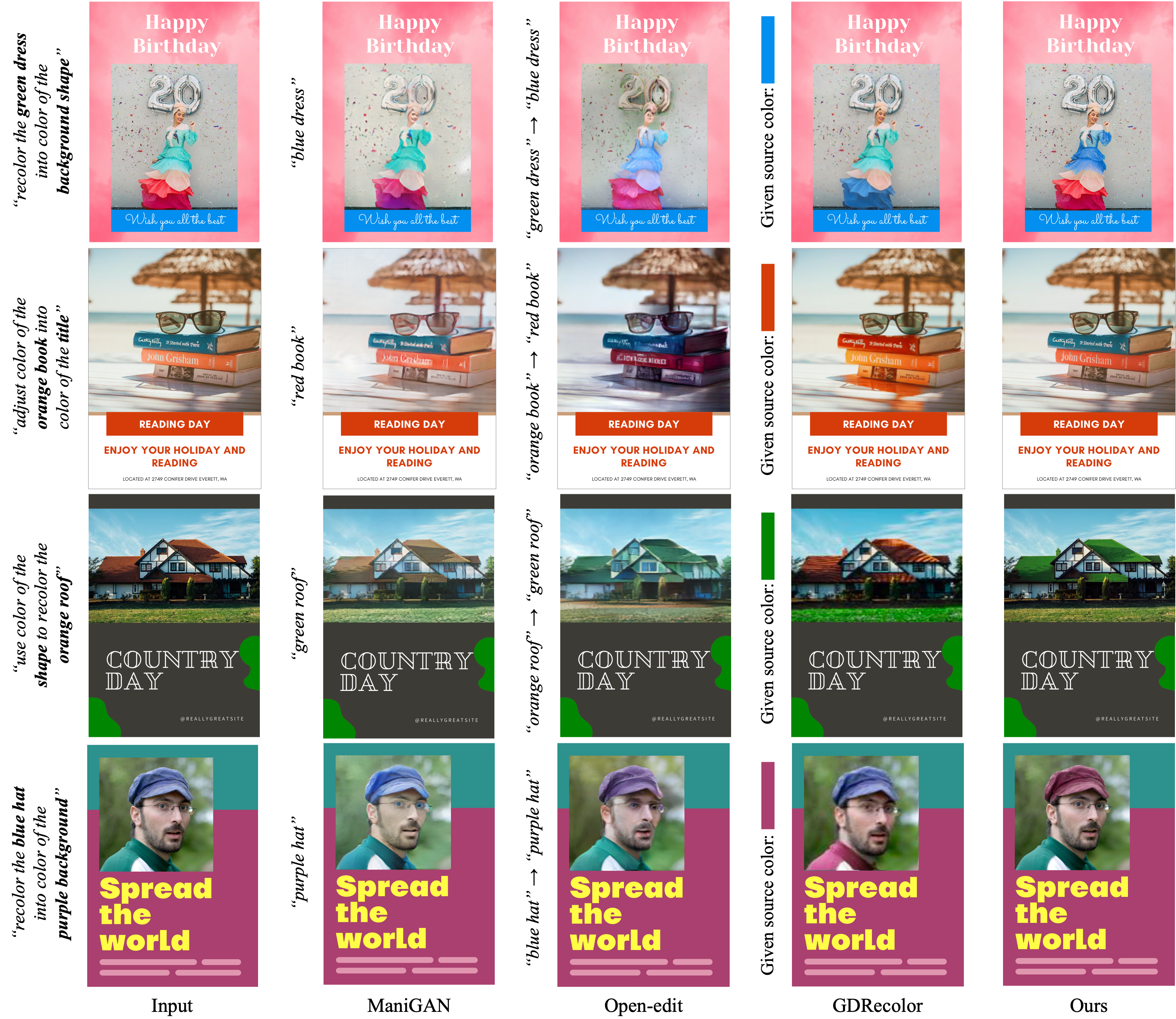}
  \caption{
  \cherry{Comparison results with state-of-the-art methods. We compare with two language-based image recoloring methods: Open-edit \cite{liu2020open} and ManiGAN \cite{li2020lightweight}; and a graphic design photo recoloring method GDRecolor~\cite{zhao2021selective}.
  }
  }
  \vspace{-2mm}
  \label{fig:qualitative_overall_results}
  
\end{figure*}

We show the qualitative results in~\reffig{qualitative_overall_results}. 
\rynn{We can see that} even though the images are recolored following the given instructions to some extent, the two language-based baselines fail to produce high-quality recoloring results. ManiGAN tends to perform global color editing (\eg~$1^{st}$ and $3^{rd}$ rows) and sometimes produces nearly unmodified results (\eg~$2^{nd}$ and $4^{th}$ rows). This is mainly because of the tradeoff between image manipulation and reconstruction during the ManiGAN training. While Open-edit can limit the editing \rynn{to the target regions locally}, it may generate unnatural contents. As the image editing is conducted in the visual-semantic embedding space, local details may be lost in the generated results (\eg~human face in the $1^{st}$ and $4^{th}$ \rynn{rows}). Besides, by using the color \rynn{words, \eg~``red'' and ``green'', their \wzww{source} colors} may not be accurate enough, failing to achieve a more visually harmonious design (\eg~hat in the $4^{th}$ row). 
Though GDRecolor can recolor in a local region, \rynn{producing more} natural and high-quality results, their method fails to follow the instructions and is less controllable. Users can only adjust \rynn{the color in the} model-predicted regions. For example, in the $3^{rd}$ row, with the green as the \wzww{source} color, the model prefers to adjust the grass color rather than the roof specified by the user. 
In contrast, our method allows flexible target region specification via instructions. Besides, our model can parse multi-granularity instructions efficiently by predicting accurate \wzww{source} colors and recoloring the image naturally, resulting in more aesthetically pleasing designs.
\revise{Due to the unavailable code and the limited space, you may refer to the comparisons with RecolGAN and Imagic in the \wzw{supplementary material}.}

\subsubsection{Quantitative Results}
\cherry{
We show the \rynn{quantitative results in~\reftab{PSNR_SSIM} on} \wzw{two} different metrics: the Peak Signal-to-Noise Ratio (PSNR) and the Structural Similarity index (SSIM). These metrics are commonly used in image editing and generation tasks~\cite{bau2020semantic,zhang2017real}. Since GDRecolor cannot follow the instruction for locating the \rynn{target} regions, the quantitative comparison with this method becomes meaningless. We thus only compare with language-based baselines in the quantitative experiment \rynn{as well as the user study to be introduced next}. The results show that our model outperforms \rynn{the} baselines by a large margin \rynn{on} \wzw{both} metrics, demonstrating that our model can generate more faithful recoloring results.}

\begin{table}
  \caption{\cherry{Recoloring results compared with language-based methods.}}
  \label{tab:PSNR_SSIM}
  \footnotesize
  \begin{tabular}{ccc}
    \toprule
    Methods & PSNR $\uparrow$ & SSIM $\uparrow$\\
    \midrule
    Open-Edit \cite{liu2020open} & 15.55 & 0.5230\\
    \midrule
    ManiGAN \cite{li2020lightweight} & 16.91 & 0.6662\\
    \midrule
    \textbf{Ours} & \textbf{24.27} & \textbf{0.8500}\\
  \bottomrule
\end{tabular}
\end{table}

\subsubsection{User Study}
\cherry{Since the metrics \rynn{used} above mainly measure the similarity between \rynn{the results and ground truth}, we further conduct a user study to validate our approach \rynn{\wzw{according to} the naturalness of the recoloring results, and the faithfulness of the results to the instructions}. We randomly select a set of design-instruction pairs (\ie~68) from our test set.}
\cherry{Participants were invited to complete a questionnaire consisting of 30 pairwise comparisons \wzww{in person, using the same color-calibrated screen.}} There are a total of 35 participants \rynn{in our study, recruited from a local Computer Science department.} The results of \rynn{this} user study are presented in ~\reftab{user_study}.
\rynn{We can see that} our model significantly outperforms existing language-based image recoloring methods in both naturalness and faithfulness. Given the challenging task of parsing language-based instructions in the graphic design context, there is still a gap between our results and those generated by designers. \rynn{More details of the user study can be found in the \wzw{supplementary material}}.

\begin{table}
  \caption{\cherry{Results of the user study.}
   We compare our method (Ours) \rynn{with} either designers or previous methods using 2AFC pairwise comparisons. \wzww{All preferences are statistically significant (p < 0.05, chi-squared test).}} 
  \label{tab:user_study}
  \footnotesize
  \begin{tabular}{lcc}
    \toprule
    vs. Methods & \multicolumn{2}{c}{\% Preferred Ours}\\
    & Naturalness & Faithfulness\\
    \midrule
    Open-Edit \cite{liu2020open} & 82.2\% & 83.0\%\\
    \midrule
    ManiGAN \cite{li2020lightweight} & 69.6\% & 82.6\%\\
    \midrule
    Designers  & 42.2\% & 36.5\%\\
  \bottomrule
\end{tabular}
\end{table}

\subsection{Effect of the Language-based \wzww{Source} Color Prediction Module}
\label{sec:experiment_color_prediction}
\cherry{To evaluate the effectiveness of the two main subnets in the language-based \wzww{source} color prediction module, we design \rynn{two variants as}:}
\begin{itemize}[leftmargin=*]
    \item \cherry{Ours without granularity recognition subnet (\textbf{Basic}). 
    We use uniform mean voting for both text elements and filled-color-based elements. Besides, we only train a single class-agnostic \wzww{source} color refinement subnet.} 
    \item \cherry{Ours without class-aware color prediction subnet (\textbf{$+$granularity}). We replace the class-aware voting strategy with the mean voting for all element types. The other components, including the granularity recognition subnet, remain the same as our final model.}
    
\end{itemize}

\subsubsection{Quantitative Results}
\begin{figure}[t]
  \centering
  \includegraphics[width=\linewidth]{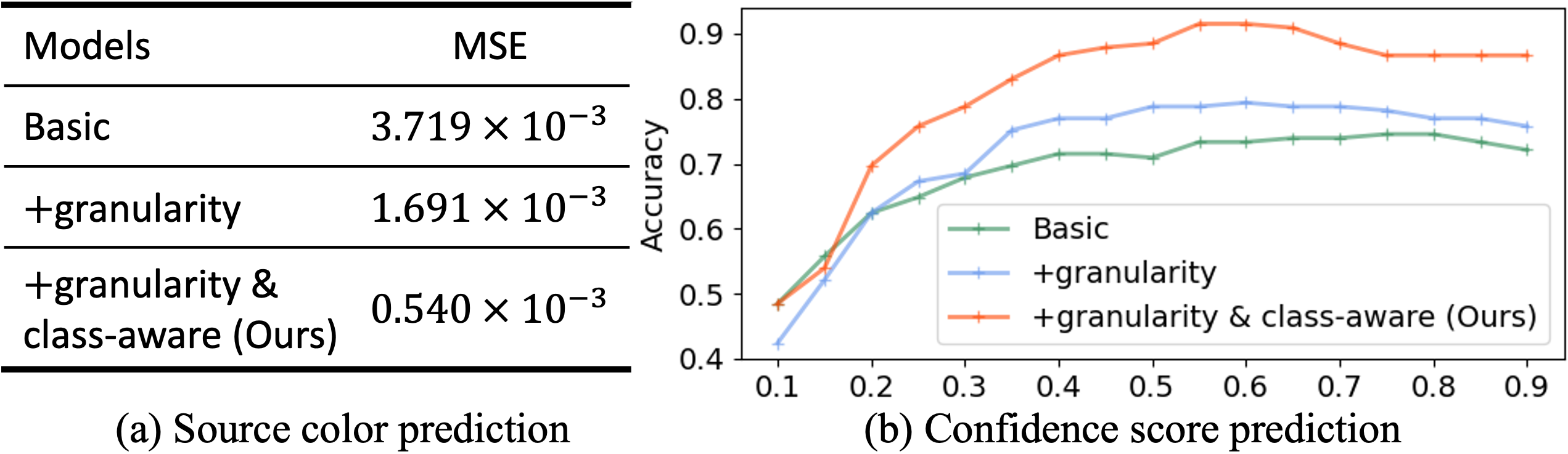}
  \caption{
  \revise{The effect of the granularity recognition subnet (granularity) and the class-aware color prediction subnet (class-aware).}}
  \label{fig:color_pred_quantitative}
\end{figure}

\cherry{
We evaluate \rynn{the predicted} \wzww{source} colors and confidence scores using different metrics.
For \wzww{source} colors, we calculate the mean square error (MSE) between \rynn{the predictions and ground truth. The results are shown}  \revise{in~\reffig{color_pred_quantitative}(a).}}
\rynn{We can see} that our complete language-based \rynn{\wzww{source}} color prediction module achieves the best performance with the smallest MSE value, showing that the \wzww{source} colors predicted by our method \rynn{are} very close to the ground truth (\ie~ an average error of $\pm 5.95$ for each of \rynn{the} RGB values $\in [0,255]$). 
\cherry{The better performance of \textit{ $+$granularity} over the \textit{Basic} variant demonstrates that jointly training with \rynn{the} granularity recognition task can enhance the model's ability on parsing design context \rynn{based on the} instructions. Besides, further \rynn{incorporating} a class-aware color prediction subnet can \rynn{help} increase the performance, leading to the best result, which is our final model.}

\revise{The confidence scores \rynn{determine} the number of \wzww{source} colors extracted based on the instructions. 
We regard the confidence prediction as a correct one if its corresponding predicted \wzww{source} color matches with a ground truth \wzww{source} color while the confidence score is larger than a threshold.
As this threshold for recognizing correctness \rynn{influences} the performance, 
we compute the accuracy under different thresholds, and show the result in~\reffig{color_pred_quantitative}(b). \rynn{We can see that
the absence of either the class-aware color prediction subnet or both subnets can significantly affect} the accuracy w.r.t the confidence scores of \rynn{the} predicted \wzww{source} colors. 
A threshold of $0.55$ for the confidence score achieves the best performance on our test set. \rynn{Refer to the \wzw{supplementary material} for more details on the evaluation of the confidence scores}.}

\subsubsection{Qualitative Results}
\begin{figure*}[t]
  \centering
  \includegraphics[width=\linewidth]{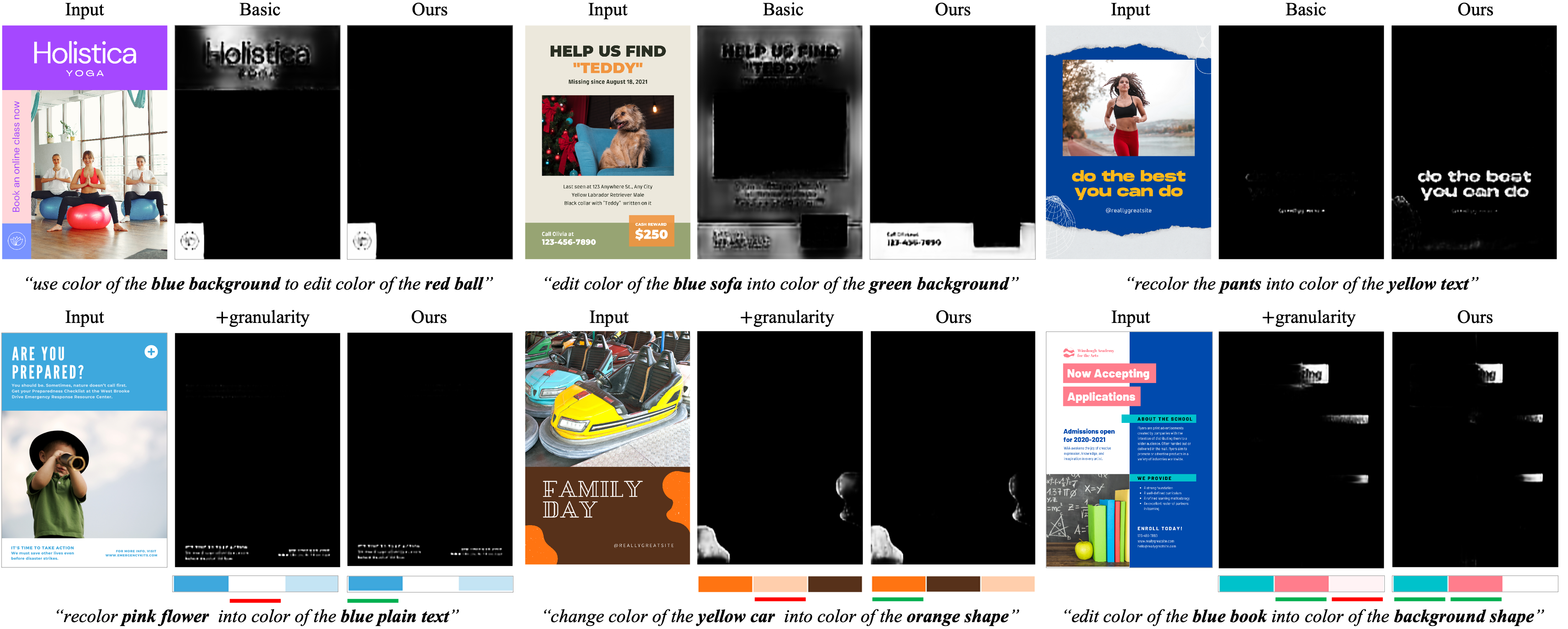}
  \caption{\cherry{What does our language-based \rynn{\wzww{source}} color prediction module learn? We compare our model with two variants: \textit{Basic} (without \rynn{granularity recognition nor} class-aware subnets in the $1^{st}$ row) and \textit{$+$ granularity} (without class-aware color prediction subnet in the $2^{nd}$ row). We show the predicted element mask in each case and predicted colors only for the examples in the $2^{nd}$ row. The bar below the color indicates its confidence score is larger than $0.5$ with green for a correct prediction and red for a wrong \rynn{one}. 
  \revise{Please zoom in for details.}
  }}
  \label{fig:color_pred_ablation}
\end{figure*}

\cherry{To further understand what the language-based \wzww{source} color prediction \rynn{module} actually learns, we qualitatively compare our method with the two variants and show the results in~\reffig{color_pred_ablation}.}

\cherry{As can be seen, without the help of the granularity recognition subnet, the model fails to distinguish between coarse-grained and fine-grained \rynn{instructions,} and locates wrong or inaccurate design elements. For example, in the $1^{st}$ row, the \textit{Basic} model fails to locate the accurate partial background region (\eg~``blue background'' and ``green background'') but predicts the whole background, and predicts the subtitle for ``yellow text'' in the $3rd$ example by mistake.
Regarding the \wzww{source} color prediction, without the class-aware color prediction \rynn{subnet}, the model can generate noisy background color (\eg~``blue plain text'' in the $2^{nd}$ row) for text elements, and noisy color in the transition region (\eg~``orange shape'' in the $2^{nd}$ row) for shape elements.
Instead, our full model can \wzw{accurately} predict \rynn{both} design element masks and \wzww{source} colors for various instructions.
}

\subsection{Effect of \rynn{the} Semantic-palette-based Photo Recoloring Module}
\label{sec:photo_recoloring_experiment}

\subsubsection{Baselines.} We compare our semantic-palette-based photo recoloring module with several baselines on image recoloring based on the given \wzww{source} colors:
    scribble-based recoloring \cite{levin2004colorization}; 
    deep user-guided recoloring \cite{zhang2017real} (with Sparse and Dense modes);
    semantic-based photo recoloring \cite{zhao2021selective} 
    \revise{(\ie, an ablation variant of our model without the semantic-palette-based region refinement);}
    Palette-based photo recoloring \zavier{\cite{tan2018efficient}}. \rynn{More details can be found in the \wzw{supplementary material}}.

\subsubsection{Qualitative Results}
\begin{figure*}[t]
  \centering
  \includegraphics[width=\linewidth]{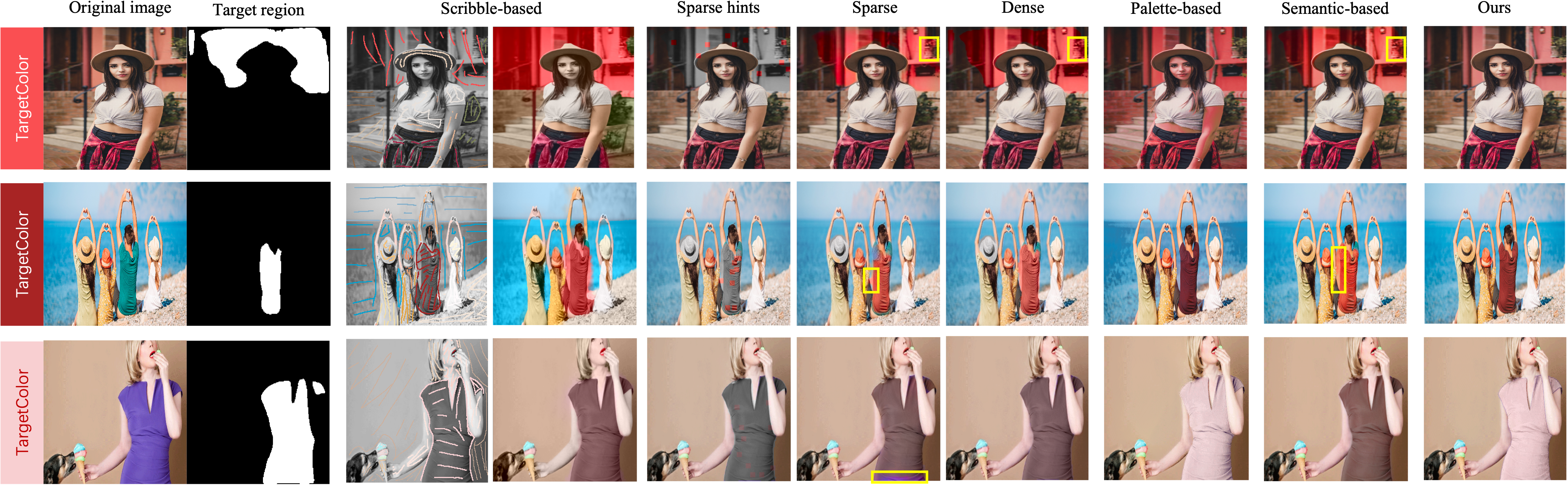}
  \caption{\cherry{Comparison with photo recoloring methods based on target regions predicted by our model. We compare with scribble-based method~\cite{levin2004colorization}, deep user-guided method~\cite{zhang2017real}, semantic-based method~\cite{zhao2021selective}, and palette-based method \cite{tan2018efficient}.
  }}
  \label{fig:qualitative_recoloring_results}
\end{figure*}

\cherry{
We show the qualitative comparison in \reffig{qualitative_recoloring_results}. Though the scribble-based, deep user-guided, and semantic-based methods~\cite{levin2004colorization,zhang2017real,zhao2021selective} can propagate the \wzww{source} color in \rynn{the} local region, they generate explicit artifacts because of the imperfect target region, such as color bleeding near the red T-shirt in the $2^{nd}$ row. Note that the scribbles and hint points of \rynn{the} compared methods~\cite{levin2004colorization,zhang2017real} may need to be carefully drawn \rynn{in order to obtain} high-quality results. Besides, the color of the resulting images may not follow the \wzww{source} color for scribble-based and semantic-based methods (\eg~\rynn{dress incorrectly colored in dark rose in the} $3^{rd}$ row). This may be because these methods rely on \rynn{Lab} color space by maintaining the \rynn{L-channel} to be the same as the input, leading to misaligned results if the \rynn{values of the L-channel are very} different between the \wzww{source} color and the original color. As for the palette-based method, without the constraint on the local region, it can generate unnatural results, such as the skin in the $1^{st}$ row. Our approach can recolor the image properly and naturally even 
the initial target region mask is not compact \rynn{nor} complete.
}

\input{sections/evaluation_usability_study.tex}

%% file: sections/evaluation_usability_study.tex
\subsection{\revise{Effect of Using the Language-based Input \wzw{Modality}}}
\label{sec:input_modality_experiment}
\begin{figure}[t]
  \centering
  \includegraphics[width=\linewidth]{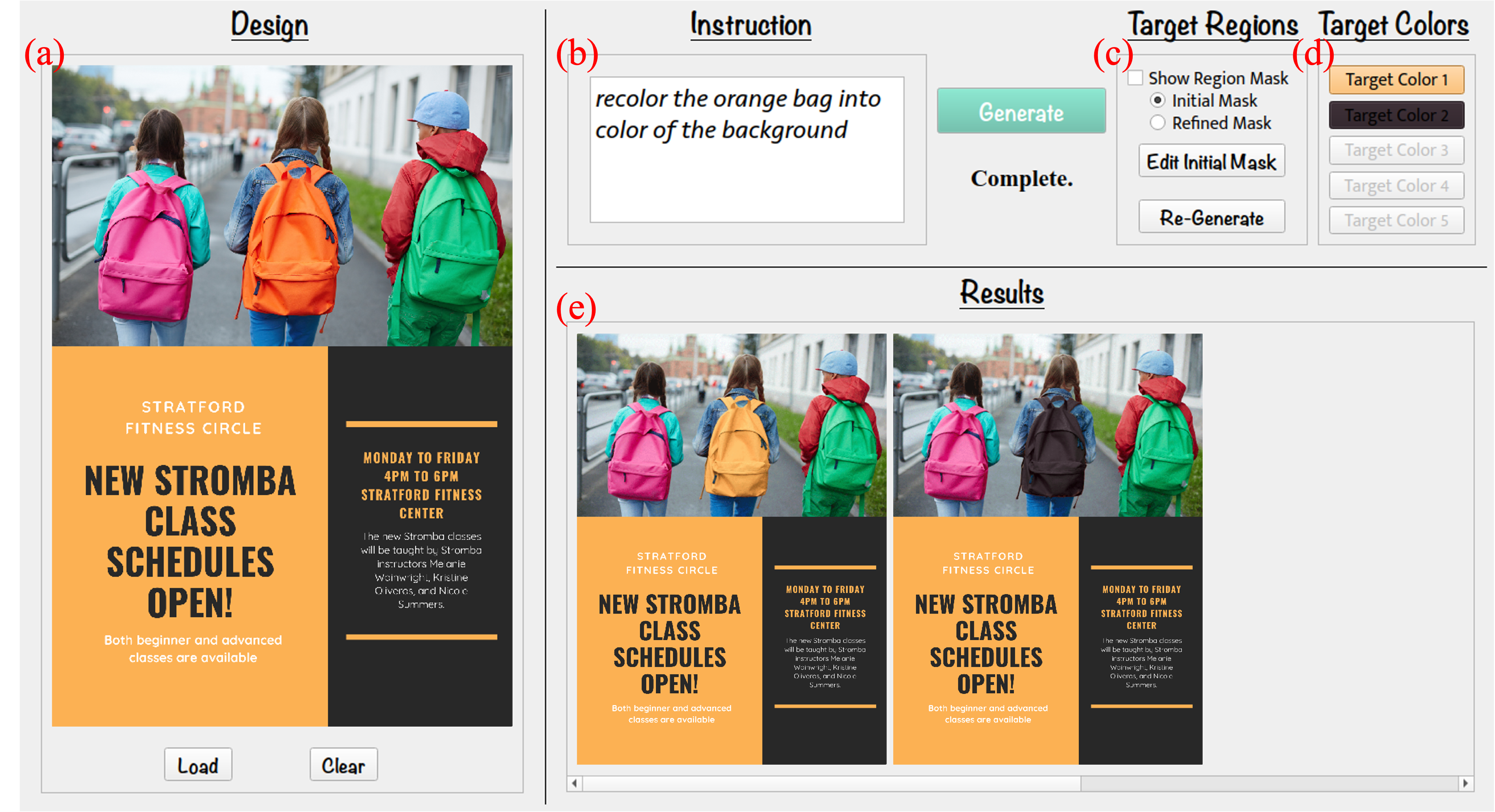}
  \caption{\revise{Our interface. (a) is the input design with a photo. (b) is the input language-based instruction. (c) and (d) display target region and \wzww{source} color predictions. (e) shows the recoloring results based on the instruction in (b).}
  }
  \label{fig:demo_interface}
\end{figure}

\begin{figure}[t]
  \centering
  \includegraphics[width=\linewidth]{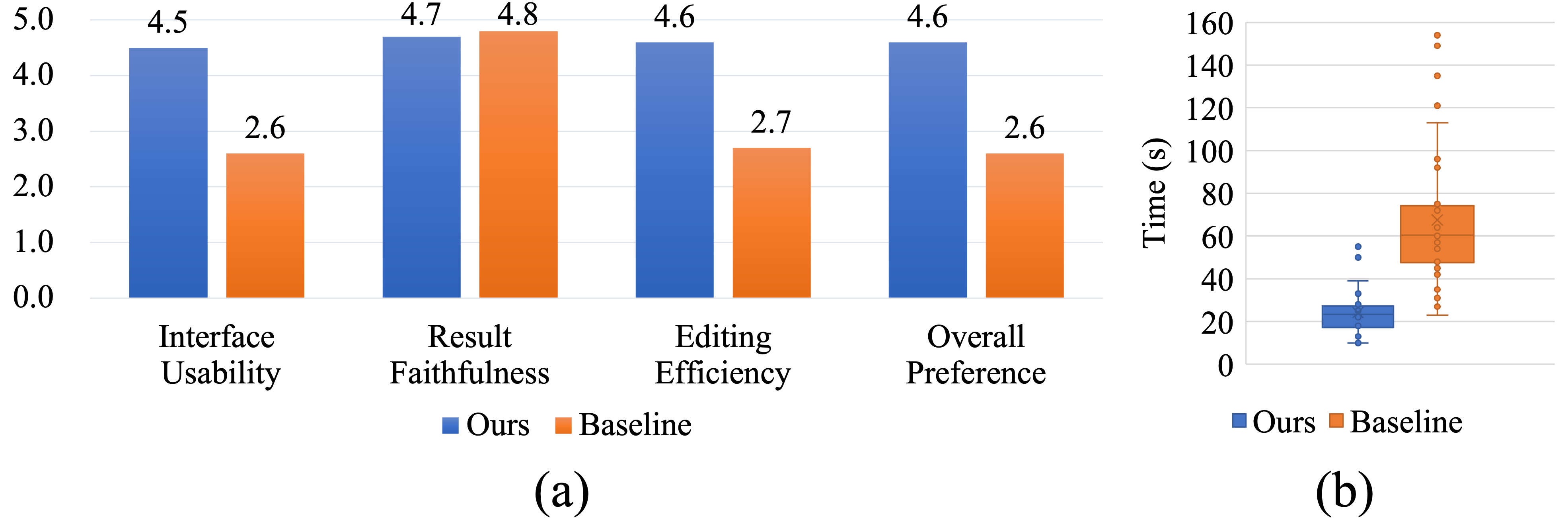}
  \caption{\revise{Results of the usability study. (a) is the users' average rating score for two interfaces from different perspectives, including usability of the interface, faithfulness of the results to the task instruction, \wzw{efficiency of finishing} an editing task, and overall preference of the interface. (b) is a box plot showing the statistics of the time usage for different design \rynn{cases}.}}
  \label{fig:usability_study_results}
\end{figure}

\revise{
We conduct a usability user study~\cite{xiao2021sketchhairsalon,o2015designscape} to further evaluate the effectiveness of our language-based input modality in \rynn{helping novices recolor} photos in the context of graphic designs. 
}

\subsubsection{Setup}
\revise{
We invited ten novice users to test two interfaces: 1) Language-based interface of our \textit{LangRecol} framework (\textbf{Ours}), as shown in~\reffig{demo_interface}, and
2) a traditional scribble-based interface without language as input (\textbf{Baseline}). \rynn{\textbf{Baseline}} uses the same interface layout \rynn{as} \textbf{Ours} but language inputs (b) is disabled. Users need to manually draw the initial rough region mask and select \wzww{source} colors via color pickers. For fair comparisons, both interfaces adopt our semantic-palette-based photo recoloring module for producing results.
Users took a 15-min tutorial before the real tasks. For each user, we randomly selected 5 design-instruction pairs from the test set as the recoloring task and asked the user to recolor following the given instructions. 
The order that users test the two interfaces is set to random, and the time usage for each design case is recorded. After conducting the design tasks, \rynn{users} were asked to provide \wzw{a} five-point System Usability Scale (SUS, 1 = strongly disagree to 5 = strongly agree) on four different perspectives each interface.
More details can be found in the \wzw{supplementary material}. }

\subsubsection{Results}
\revise{The average rating scores for different perspectives and the statistics of \rynn{interaction} time are shown in~\reffig{usability_study_results}. As can be seen, our interface is rated significantly higher than the baseline interface for most perspectives except for faithfulness, as the fully human-guided results should be most consistent with human intentions. For the \rynn{interaction} time, our language-based input can help save 43s ($63.70\%$ of the time) per example on average, which indicates the effectiveness of our language-based interface. 
Users also provided many positive feedback on our interface, such as ``This interface is really cool as I can create diverse (recoloring) designs with only a single click'' and ``I enjoy this creative interface, where I can directly tell the system what I want via typing or even voice in the future''. \rynn{Additional user feedback can be found in the \wzw{supplementary material}}.}

%% file: sections/application.tex
\section{Applications}
\cherry{We depict several practical applications enabled by our framework.}

\begin{figure}[t]
  \centering
  \includegraphics[width=\linewidth]{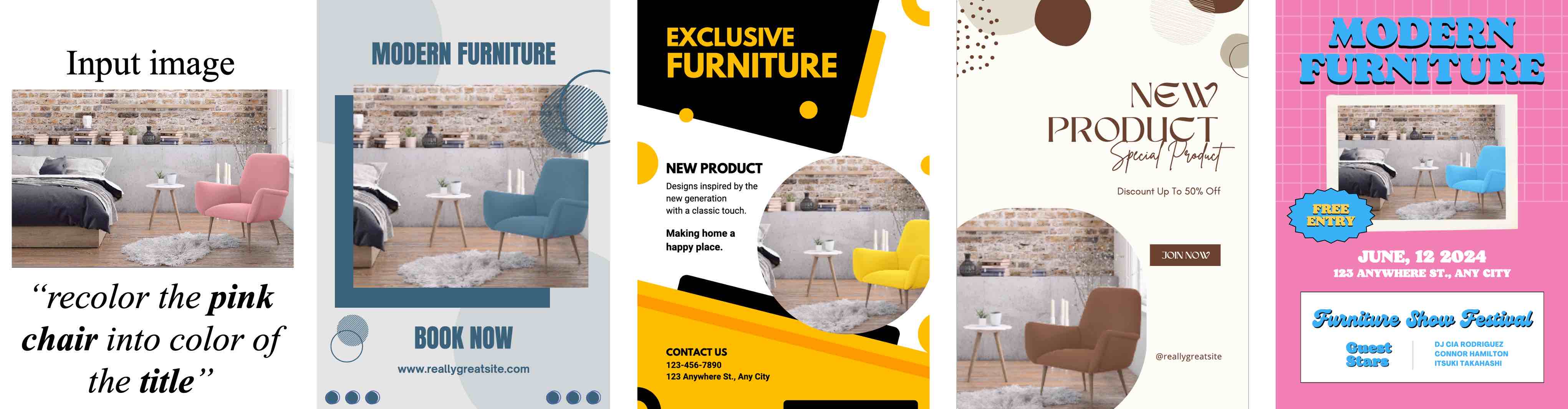}
  \caption{ \cherry{Design template pairing. Our model can adjust photo color based on a set of design templates for quick brainstorm and authoring.}
  }
  \label{fig:application1}
\end{figure}

\textbf{Design Template Pairing.}
\cherry{
As the graphic design community grows, there are many well-curated online design template repositories. It is quite often that users need to find a good matching template for their given photo, such as an event poster. However, directly adjusting the photo color based on each individual template is tedious and it may take a long time to find the proper one. With our model, users can examine a set of templates by automatically recoloring the image to match \rynn{with each of the templates} with a single instruction, as shown in~\reffig{application1}. This allows designers to brainstorm ideas, and novices to finalize their template in an efficient way.}

\begin{figure}[t]
  \centering
  \includegraphics[width=\linewidth]{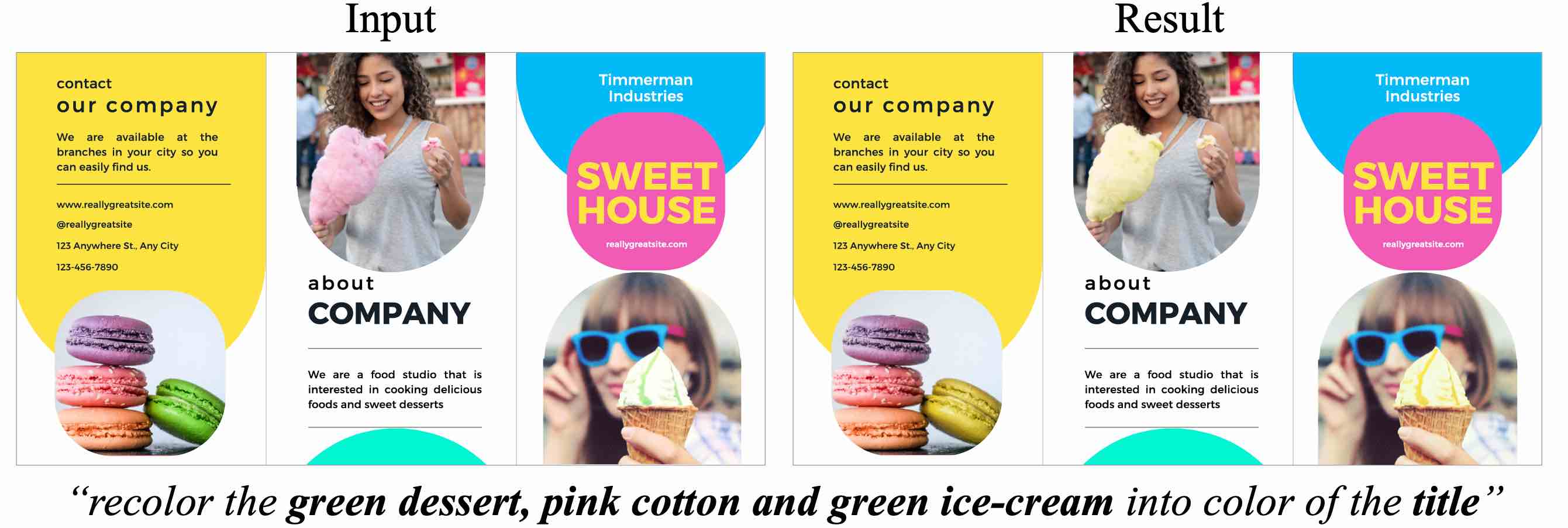}
  \caption{\cherry{Brochure Photo Recoloring. Our model can recolor multiple photos with a single design.}
  }
  \label{fig:application_2}
\end{figure}

\textbf{Brochure Photo Recoloring.}
\cherry{Some types of graphic design may contain more than one photo, \rynn{\eg~brochure}. Our model can also deal with \rynn{this case} by recoloring multiple inserted photos together with a single instruction. As the photos may not share the same semantic content, we adapt instructions containing multiple target \wzw{region descriptions} by separating the instruction into parts including only a single target region \wzw{description}.
\rynn{\reffig{application_2} shows some example results.
We leave more complex instructions to future work.}} 

\begin{figure}[t]
  \centering
  \includegraphics[width=\linewidth]{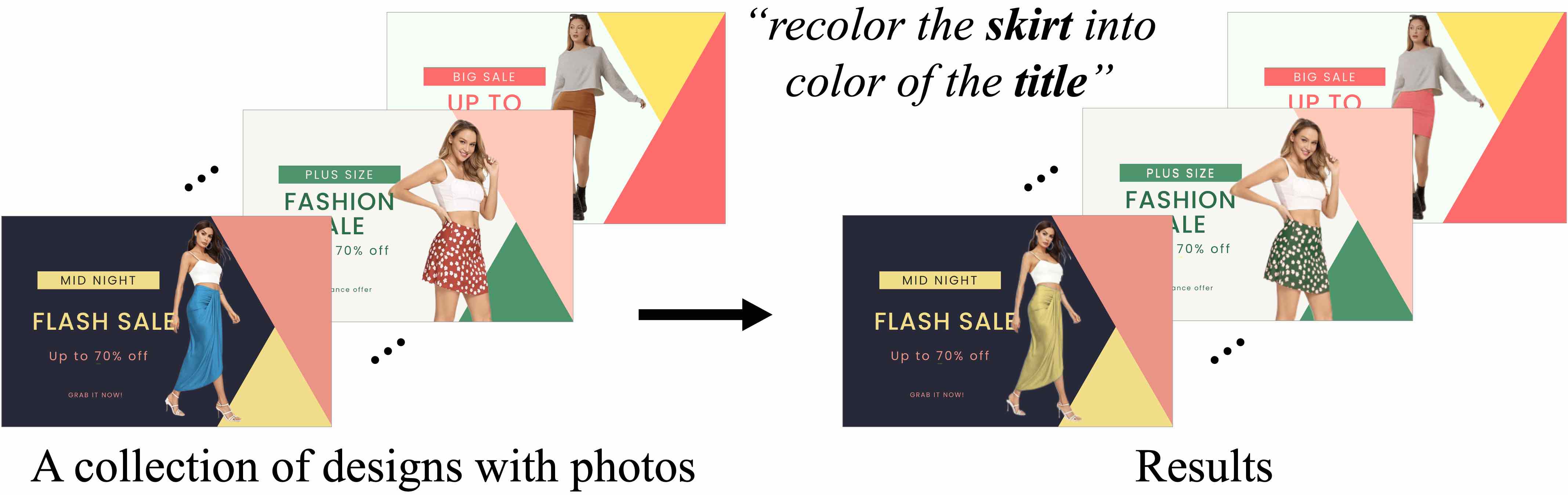}
  \caption{\cherry{Recoloring a design collection. We show a design case for banners used in online promotion.}
  }
  \label{fig:application_3}
\end{figure}

\textbf{Recoloring a Design Collection.}
\cherry{Managing an online business (\ie~e-commerce), such as a clothing shop on Amazon, becomes more and more popular, especially \rynn{after the} Coronavirus pandemic. People (\eg~businessmen) need to produce banners and advertisements efficiently to incorporate the fast-changing trend. As shown in \reffig{application_3}, with a single instruction, our model enables recoloring for a design collection containing different images.}

\begin{figure}[t]
  \centering
  \includegraphics[width=\linewidth]{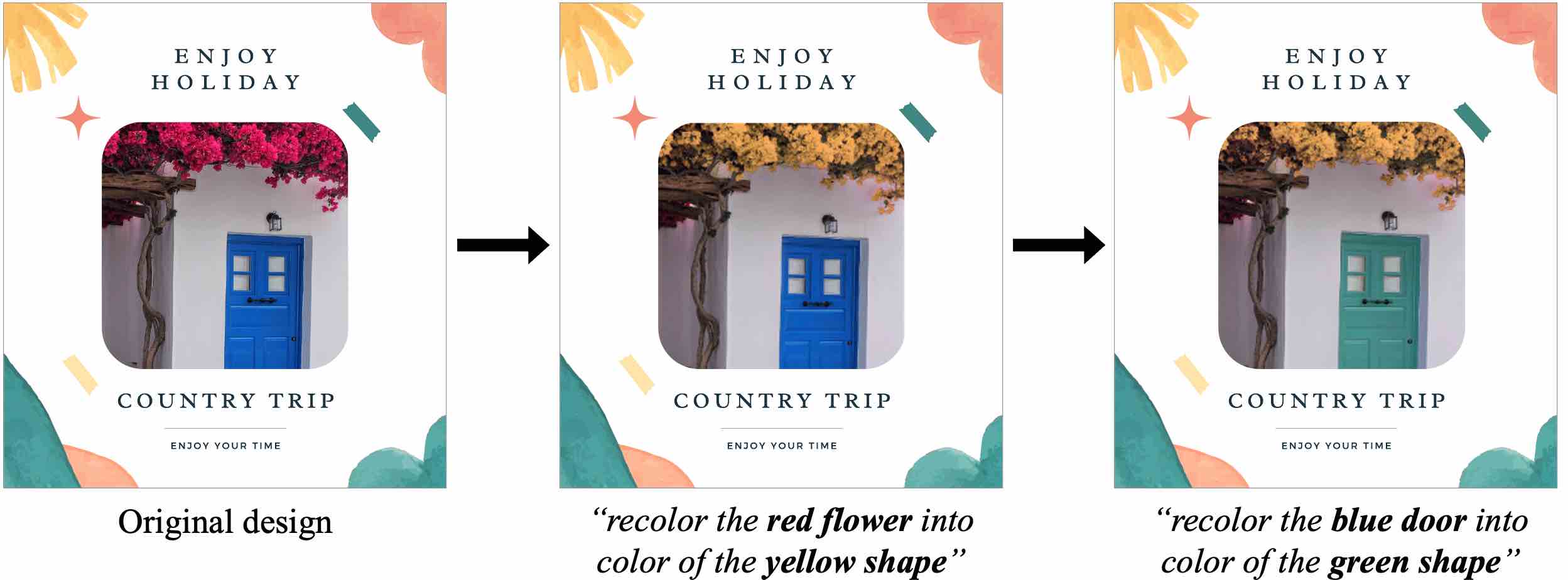}
  \caption{ \cherry{Iterative graphic design photo recoloring.}
  }
  \label{fig:application_4}
\end{figure}

\textbf{Iterative Graphic Design Photo Recoloring.}
\cherry{Iterative design is a common strategy in general editing tasks. Our model also provides this function, where users can continually adjust the photo color until reaching \wzw{satisfactory} results, as shown in  \reffig{application_4}.}

%% file: sections/conclusion.tex
\section{Conclusions}
\cherry{In this work, we \rynn{have introduced} a new language-based method for photo color adjustment in graphic \rynn{designs}. With multi-granularity instructions, users can express their design intention freely. Our method includes two essential components: the language-based \wzww{source} color prediction module and the semantic-palette-based photo recoloring module. A language-based design synthesis method is proposed to enable the training of our framework. \rynn{We have validated the effectiveness of our model through comprehensive experiments. Our model outperforms existing methods both quantitatively and qualitatively. We have also shown a \rynn{number} of practical applications of our model. We will release our code and dataset to encourage} more research on language-based graphic design editing.}

\begin{figure}[t]
  \centering
  \includegraphics[width=\linewidth]{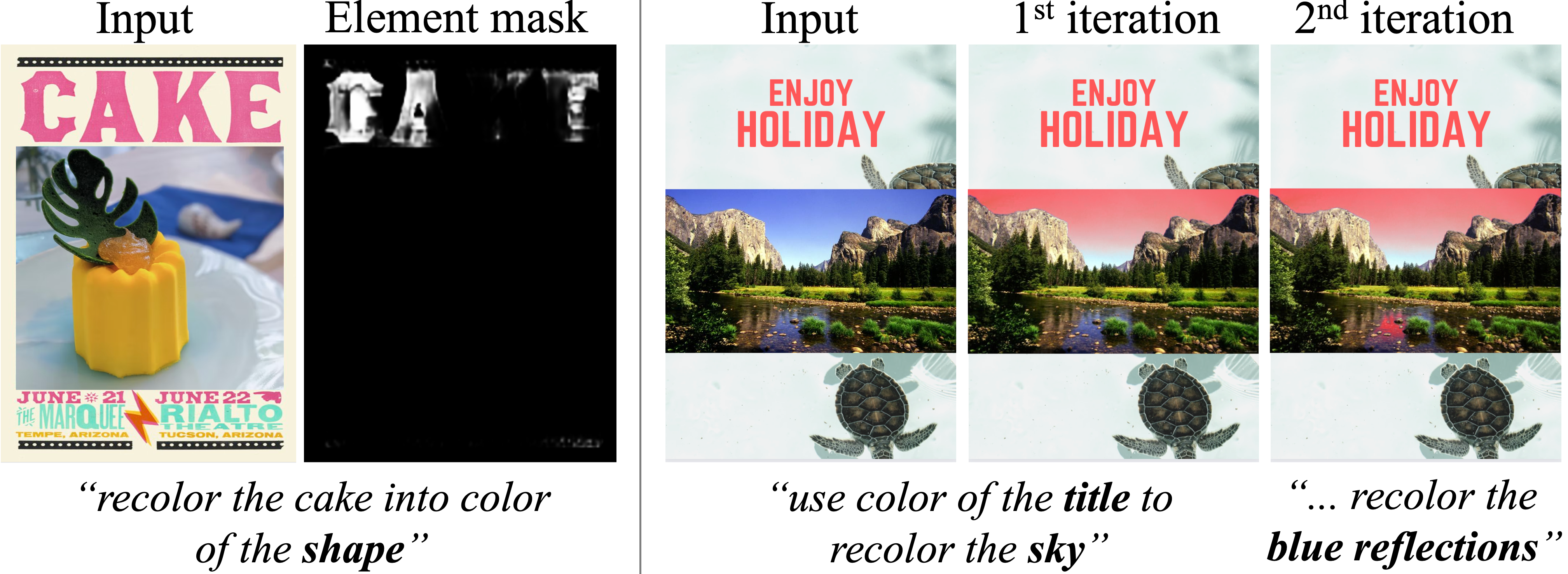}
  \caption{\cherry{Failure cases.}}
  \label{fig:failure_cases}
\end{figure}

\textbf{Discussions.} Our \rynn{work opens up} a new venue for adjusting photo color in graphic \rynn{designs} with language-based instructions, but it still has \wzww{limitations} and room for further improvement. First,  \wzww{following a state-of-the-art referring image segmentation work \cite{feng2021encoder}}, the language model we use here is a basic Bi-GRU trained on our dataset from scratch, and it may not work if the target object or design element \rynn{does not appear in our training set}. A possible solution is to utilize a \rynn{more} powerful pretrained cross-modality model, \eg~CLIP~\cite{radford2021learning}. \wzww{However, simply replacing the text encoder may not significantly improve this limitation since our language-based instructions are unique (\ie~with two parts of information referring to source color and target region), and are different from the prompts used for training existing cross-modality model. Nevertheless, it would be interesting to try them as a future work. } Second, even though we have tried to synthesize graphic designs with appearances as diverse as possible, we still cannot cover all the cases that \rynn{appear} in real life, especially creative design. For example, in \reffig{failure_cases}~left, since the title ``CAKE'' occupies a large portion of the design with a creative font face, our model misinterprets it as a combination of shapes. We \rynn{may} consider alleviating this problem in the future by finetuning our language-based \wzww{source} color prediction module with a dataset containing real graphic designs. Third, our model may have difficulties in preserving long-range consistency, such as \rynn{reflection of the sky in \reffig{failure_cases}~right}. To address it, further instructions can be added to iteratively refine the result, such as using ``... recolor the blue reflections ...'' in the example. 
\wzww{More discussions on the scope and potential ethics issues of our work can be found in the supplementary material.}

\begin{acks}
We thank the anonymous reviewers for the insightful comments and constructive suggestions on our paper. This work is in part supported by a GRF grant from the Research Grants Council of Hong Kong (Ref. No.: 11205620).
\end{acks}